\documentclass[
]{ceurart}

\usepackage{listings}
\usepackage{float}
\usepackage{siunitx}
\usepackage{graphicx}
\begin{document}

\copyrightyear{2024}
\copyrightclause{Copyright for this paper by its authors.
  Use permitted under Creative Commons License Attribution 4.0
  International (CC BY 4.0).}

\conference{HAII5.0: Embracing Human-Aware AI in Industry 5.0, at ECAI2024, 19 October 2024, Santiago de Compostela, Spain.}

\title{Towards Differentiating Between Failures and Domain Shifts in Industrial Data Streams} 

\author[]{Natalia Wojak-Strzelecka}[%
orcid=0009-0003-6161-0697,
email=nataliawojak3@gmail.com,
]
\cormark[1]
\fnmark[1]
\address[1]{Jagiellonian Human-Centered AI Lab, Mark Kac Center for Complex Systems Research, Institute of Applied Computer Science, Faculty of Physics, Astronomy and Applied Computer Science, Jagiellonian University, ul. prof. Stanisława Łojasiewicza 11, 30-348 Krakow, Poland}
\address[2]{Institute of Computing Science, Poznań University of Technology, 60-965 Poznań, Poland}

\author[1]{Szymon Bobek}[%
orcid=0000-0002-6350-8405,
email=szymon.bobek@uj.edu.pl,
]
\fnmark[1]

\author[1]{Grzegorz J. Nalepa}[%
orcid=0000-0002-8182-4225,
email=grzegorz.j.nalepa@uj.edu.pl,
]
\fnmark[1]

\author[2]{Jerzy Stefanowski}[%
orcid=0000-0002-4949-8271,
email=jerzy.stefanowski@cs.put.poznan.pl,
]
\fnmark[1]

\cortext[1]{Corresponding author.}
\fntext[1]{These authors contributed equally.}

\begin{abstract}
Anomaly and failure detection methods are crucial in identifying deviations from normal system operational conditions, which allows for actions to be taken in advance, usually preventing more serious damages.
Long-lasting deviations indicate failures, while sudden, isolated changes in the data indicate anomalies.
However, in many practical applications, changes in the data do not always represent abnormal system states. 
Such changes may be recognized incorrectly as failures, while being a normal evolution of the system, e.g. referring to characteristics of starting the processing of a new product, i.e. realizing a domain shift. 
Therefore, distinguishing between failures and such ''healthy'' changes in data distribution is critical to ensure the practical robustness of the system.
In this paper, we propose a method that not only detects changes in the data distribution and anomalies but also allows us to distinguish between failures and normal domain shifts inherent to a given process. 
The proposed method consists of a modified Page-Hinkley changepoint detector for identification of the domain shift and possible failures and supervised domain-adaptation-based algorithms for fast, online anomaly detection.
These two are coupled with an explainable artificial intelligence (XAI) component that aims at helping the human operator to finally differentiate between domain shifts and failures.
The method is illustrated by an experiment on a data stream from the steel factory.
\end{abstract}


\maketitle

\section{Introduction}


One of the most common approaches for automated monitoring of industrial processes is anomaly and failure detection mechanisms.
They allow for unsupervised health state monitoring of an industrial facility and can reduce costs related to maintenance and unplanned stoppage in production by alerting the operators of upcoming detected problems.
However, in many practical applications, industrial processes frequently adapt to shifts in production lines or undergo on-demand reconfiguration, resulting in data patterns that may superficially appear as failures while being a healthy evolution of the system or a consequence of variations in products' specifications.
This may cause anomaly detection or failure detection systems to trigger many false positive alarms in case of technical changes in the production line, or in case of nontypical production line configuration.
Our research aims at differentiating between healthy changes in the data distribution and genuine failures that in such a scenario are indistinguishable from state-of-the-art approaches~\cite{10.1007/978-3-031-36027-5_37}.
By healthy change, we understand the permanent (in a certain time span) change in data distribution that is not the result of a failure but is a consequence of a characteristic of the process that the data comes from. 

In this paper, we focus on the dataset from the cold rolling facility of the steel factory. 
The cold rolling process aims to reduce the steel thickness to a satisfactory level, making the product ready to be sold to customers.
The dataset was generated with the simulator that reflects the physical processes of the real facility\footnote{The generator was developed in collaboration with a steel factory in Poland, and its source code cannot be released. However, the dataset used in our work, along with the source code for the experiments to ensure full reproducibility, is available at \href{https://github.com/nataliawojak/ECAI-workshop-2024}{GitHub}}. Such data present the state of many sensors in the production line as a function of time and can be treated as a data stream~\cite{stefanowski2017stream}.
In such a scenario, the following challenges can appear:
\begin{itemize}
\item Domain shifts --  The facility produces a variety types of products, which differ in their chemical properties and target thickness. This requires the production line to adjust rolling parameters, which in the data stream may appear as domain shift, often falsely detected as failure. In general, the domain shift represents a change in the data characteristics (e.g. change in data distribution that is caused by different production parameters for different products).
\item Anomalies -- Short-lasting deviations in product thickness that can make the product defective and not be sold. 
\item Failures -- Very rarely the failures in the facility components such as bearings, engines, rolls, etc. may cause larger problems with the facilities. From a purely detection perspective, they may not be distinguishable from the domain shifts, yet carry serious consequences when not spotted fast enough.
\end{itemize}

Most of the earlier work has been focused on solving only one of these challenges.
However, in many practical applications, a comprehensive approach is needed that addresses all three cases simultaneously. 
This is important, especially in situations where many domain shifts are present in the data stream.
In such a scenario, healthy but different from the statistical point of view data (e.g. new products, rare products, equipment upgrades, etc.) can be mistakenly interpreted by the automated algorithms as failures, causing a lot of false positive alarms, which leads to stoppages of production and generating large, unnecessary maintenance costs. 
It happens because failures and domain shifts usually are visible as the change in the data distribution and therefore may be indistinguishable for automated 
detection methods.
This makes the problem even more challenging in situations where domain adaptation algorithms are used to quickly update the model between healthy domain shifts~\cite{ganin2016domainadversarial}.
The domain adaptation techniques allow learning a model from a source data distribution and updating that model on a different (but related) target data distribution (e.g. distributions isolated after the domain shift).
In cases where there is no differentiation between failure and domain shift, the domain adaptation mechanism can falsely consider a failure as a healthy domain shift and cause serious damage by seamlessly adapting the model to it.

On the other hand, it is not feasible for human experts to investigate and analyze each of the cases of possible failure/domain shift manually.
We argue that developing new techniques to identify potential domain shifts and failures in the data, along with the assistance of a human operator supported by explainable artificial intelligence tools, can improve the online health monitoring of industrial processes.

\begin{figure}[h]
\centering
\includegraphics[scale=0.71]{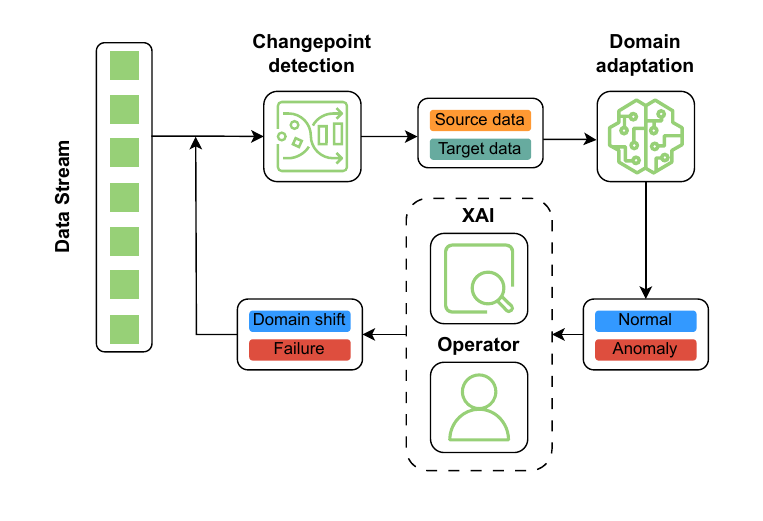}
\caption{The flow chart for differentiating between failures and domain
shifts. The changepoint detection algorithm marks possible domain shifts and failures.
After that, the domain adaptation algorithm updates the model that detects anomalies. Simultaneously human operator accompanied by the XAI algorithm decides if the change in the data represents a healthy domain shift or a failure.} \label{fig:flow}
\end{figure}

Therefore, our goal is to propose a method that not only detects domain shifts and failures but also allows to distinguish one from another.
Our method is composed of the following elements: the modified Page-Hinkley changepoint detector that identifies domain shifts and possible failures, supervised domain-adaptation-based algorithms for fast, online anomaly detection, and an explainable artificial intelligence (XAI) component that aims at helping the operator differentiate between domain shifts and failures.
The explanations provide insight into how the machine learning model uses particular sensor data between domain shifts to predict anomalies.
In a case where there is a visible change in the way the data is used by the model (e.g.similar feature value contributes very differently to the model prediction across two domain shifts), the expert may undergo additional analysis and order a stoppage of the system in a case of discovered failure or let the system operate if the change was a healthy domain shift.
The workflow for our approach is depicted in Fig.~\ref{fig:flow}.

The remainder of the paper is organized as follows.
In Section~\ref{sec:sota}, we describe the background of our research. 
In Section~\ref{sec:method} we provide a detailed description of a proposed method, which we later demonstrate on the steel factory data in Section~\ref{sec:exp}.
We summarize and discuss future extensions of this work in Section~\ref{sec:summary}.

\section{Related works}
\label{sec:sota}

In this section, we review the relevant literature and discuss previous work related to explainable artificial intelligence, anomaly detection, and data streams, along with their applications in the industry.

Anomaly detection involves identifying patterns in the data that deviate from expected or normal behavior. 
Anomalies in data often provide significant, and sometimes critical, actionable insights across a wide range of application domains, which makes detecting them crucial. For example, abnormalities in credit card transactions could signal credit card fraud or identity theft~\cite{Aleskerov1997CARDWATCHAN}. Abnormal readings from a spacecraft sensor could indicate a fault in one of the spacecraft components~\cite{Fujimaki2005AnAT}. One of the key beneficiaries of early anomaly detection is industrial companies, as identifying abnormal conditions can prevent unplanned downtime, reduced production quality, and safety breaches. 
The most commonly used algorithms include models based on neural networks such as autoencoders (AE)~\cite{Rumelhart1986LearningIR} where the goal is to learn the internal structure of the data to enable input reconstruction. 
This is useful for anomaly detection, as an autoencoder trained on normal data can reconstruct these inputs well, but struggles with anomalies, having not learned their structure during training. The other very popular approach is a 
One--class support vector machines (OCVSM)~\cite{SHIN2005395}, which identifies the characteristics of a normal class, and any observation that deviates from this learned pattern is considered anomalous. 
Density--based anomaly detection techniques, such as Local Outlier Factor (LOF)~\cite{10.1145/335191.335388}, assess the density of each data instance neighborhood. 
Instances located in low-density neighborhoods are labelled anomalous, while those in high--density neighbourhoods are considered normal. A comprehensive comparison of anomaly detection algorithms, evaluated on three benchmark datasets and industry platforms, was conducted in~\cite{9892939}. In~\cite{9564228} the authors employed a modified autoencoder to not only detect anomalous behavior in the hot rolling mill process but also identify the origins of most anomalies detected by the deep learning model.
For more details, the reader is referred to a comprehensive overview of anomaly detection techniques, covering various research areas and application domains, which can be found in~\cite{10.1145/1541880.1541882}.

Building a robust anomaly detection algorithm may not always meet all the needs of the business. 
In complex problems, understanding the factors that contribute to anomalies is crucial for gaining insights into the process. 
This understanding enables the implementation of effective maintenance actions and strategies to prevent future abnormalities. 
Therefore, employing explainable artificial intelligence solutions can be also very beneficial. 
XAI methods aim to improve understandability and transparency in the model decision-making process, highlighting which measurements contribute the most to current model decisions. 
A comprehensive study of interpretable artificial intelligence can be found in~\cite{molnar2020interpretable} or \cite{guidotti2018survey}.  
Related examples of widely used XAI methods include the following: LIME~\cite{ribeiro2016why}, SHAP~\cite{NIPS2017_8a20a862} - as methods indicating feature importance for predictions, LUX~\cite{10.1007/978-3-030-77980-1_34} or counterfactual explanations~\cite{wachter2018counterfactual} - which show how to change the input of the model to obtain a desired model output \cite{stepka2024multi}.  Nevertheless, most XAI methods are designed for static data and do not directly handle the dynamic nature of industrial data and the data shift. 
Some recent works for such applications are as follows.
The authors in~\cite{10.1007/978-3-031-50396-2_5} proposed a framework for online anomaly detection in a cold rolling mill that includes the XAI module, helping the operator better understand the origin of the anomaly and take the necessary actions. 
Ding et al.~\cite{DING2023144} applied the SHAP module to enable transparent decision--making processes within the diagnostic framework for hot-rolled strip crown.

Generating data continuously at regular time moments and observing changes in data distribution over time is a common industry practice. 
Furthermore, because the processing must be done in real--time, data stream learning~\cite{article} is a suitable method for this environment. 
Let us recall that a data stream can be defined as an unbounded sequence of instances which arrive at high speed and require fast processing \cite{stefanowski2017stream}. It introduces unprecedented challenges, especially with respect to computational resources and limited prediction time. Besides new processing requirements, another important challenge is that algorithms learning from streams often act in dynamic non-stationary environments, where the data and target concepts change over time in a phenomenon called \textit{concept drift} \cite{krawczyk2017ensemble}.

The detailed description of handling streaming data from steel industry facilities can be found in~\cite{jakubowski2024artificial}. 
Changes in the data stream (referring to concept drifts) can significantly impact model performance, potentially resulting in a decrease in production quality. 
Therefore, monitoring these changes and responding to them are essential aspects of daily operations. To address this challenge, drift detection methods could be applied, such as the Early Drift Detection Method (EDDM)~\cite{BaenaGarc2005EarlyDD}, Adaptive Windowing (ADWIN)~\cite{bifet2007learning}, Page--Hinkley (PH)~\cite{10.1093/biomet/41.1-2.100} or others~\cite{AGRAHARI20229523}.

However, the aforementioned methods do not cover all possible problems that occur in the industry environment. 
On the one hand, re-configuring production or starting processing a new product might result in incoming samples being marked as anomalies or failures. 
On the other hand, the data characteristics of new products and potential failures can be similar, making them difficult to distinguish for drift detection methods.  
In our work, we introduce a method designed to identify changepoints in data streams indicating possible data shifts and failures and provide explanations to assist operators in better differentiating between them.

\section{Method for differentiating between failures and domain shifts}
\label{sec:method}

The proposed method consists of three subsequent steps: 1) distribution change discovery, 2) domain adaptation and 3) explanation of the differences between the source model and the adapted one.
First, the detector for identification of the changes in the multivariate data stream is presented. 
These changes trigger a domain adaptation algorithm that adjusts the anomaly detection model to the changing data distribution.
Up to this point, it is still unknown if the distribution change is a healthy data shift or a failure.
Finally, an explainable artificial intelligence algorithm is used to provide the feature importance for the classification algorithm before and after the adaptation.
These differences in importance are used as a decision support component that helps the human operator differentiate between failures and domain shifts.
The general workflow of the method is presented in Figure \ref{fig:flow}.

\subsection{Changepoint detector}
Let us assume that the data instances arrive as the data stream $S = \{\mathbf{x}^{i}\}_{i=0}^{N}$, where $\mathbf{x}^{i} = [x_{j}^{i}]_{j=1}^{n}$ is $n$--dimensional attribute vector. 
The first $m$ elements of the stream $S$ constitute the reference distribution $X_{ref} = \{\mathbf{x}^{i}\}_{i=0}^{m}$. The next object in the stream will be approximate distribution $X_{approx} = \{\mathbf{x}^{m+1}\}$. 
The KL divergence between $X_{ref}$ and $X_{approx}$ is then estimated 
following the formula presented in~\cite{4595271}: 
\begin{equation}\label{eq:KL}
\widehat{D}_{KL}(X_{ref} \textrm{ }||\textrm{ } X_{approx}) = \frac{n}{m} \sum_{i=0}^{m} \log\frac{r_k(\mathbf{x}^{i})}{s_k(\mathbf{x}^{i})} + \log \frac{1}{m-1},
\end{equation}
where $r_k(\mathbf{x}^{i}) \textrm{ and } s_k(\mathbf{x}^{i})$ are respectively, the Euclidean distances to the $k^{th}$ nearest--neighbour of $\mathbf{x}^{i}$ in $X_{ref} \setminus \{x_i\}$ and $X_{approx}$. Further, $\widehat{D}_{KL}$ is added to Page--Hinkley (PH) drift detector.
If PH does not trigger an alarm (the threshold for $\widehat{D}_{KL}$ is not exceeded), the reference distribution increases by the sample $x^{m}$. Otherwise, the detected drift indicates the changepoint in data stream $S$, which brings about the need for adjustment of the reference distribution. In this case, the reference distribution consists of data stream objects starting from sample $x^{m}$ \big($X_{ref} = \{\mathbf{x}^{m}, \mathbf{x}^{m+1}, \ldots\}$\big). 

\subsection{Domain adaptation classifier}

The subsequent phase is the construction of a domain adaptation classifier $f : S \rightarrow \{0,1\}$, where 0 represents the normal instance and 1 refers to the anomaly. As a domain adaptation model, the feature--based classification and contrastive semantic alignment (CCSA) was selected~\cite{motiian2017unified}. 
We assume that the data preceding the initial changepoint constitutes a \textit{source domain} $X_s$ (in the industry scenario analyzed, the source domain data originates from the most prevalent rolled product). Let $D_s = \{(\mathbf{x}^{i}, y^{i})\}_{i=0}^{s}$, where $\mathbf{x}^{i} \in S$ and $y^{i} \in \{0,1\}$. The data coming after the changepoint belongs to the \textit{target domain} $D_t$.
To train the domain adaptation classifier, a small batch of samples is taken. Let the $batch = k$, and $X_{train} = \{(\mathbf{x}^{i}, y^{i})\}_{i=l}^{l+k}$, and $l>s$, $k \ll s$. The prediction is made for the next batch of samples $X_{test} = \{(\mathbf{x}^{i})\}_{i=l+k}^{l+2k}$. The motivation behind this step is to obtain a classifier making accurate predictions with a limited number of samples from the target domain. After prediction, the target domain is expanded with the predicted batch. This procedure continues until a new changepoint is identified. Taking into account the sudden nature of process changes, the domain adaptation model discards previous targets and initiates adaptation to the new product from the beginning.

\subsection{Explanation of the changes of model behavior between changepoints}
The last step of the proposed method focuses on explaining the decision of the domain adaptation model compared to the source model.
It allows us to observe possible abnormal changes in the usage of feature values that may indicate adaptation to failure.
We have decided to use the SHAP (SHapley Additive exPlanations)~\cite{NIPS2017_8a20a862}  because it allows for monitoring changes in feature importance within model predictions which adapts to different domains. 
We hypothesize that monitoring the evolving impact of different features on anomaly prediction over time could help in distinguishing between failure and domain shift. 

To accomplish this, the explanations are determined on each training batch of samples $X_{train}$. 
The Shapley value is determined by evaluating the feature value across all possible combinations with other features, weighting and summing the results:
\begin{equation}\label{eq:shap1}
\Phi_{j}=\sum_{K\subseteq\{1,...,n\}/\{j\}}\frac{|K|!(n-|K|-1)!}{n!}(val_x(K\cup{j})-val_x(K)),
\end{equation}
\begin{equation}\label{eq:shap2}
val_{x}(K)=\int f(x_{1},...,x_{n})d\mathbb{P}_{x\notin K}-E_{X}(f(X)),
\end{equation}
where $K$ is the subset of features in the model, and $n$ is the total number of features, $x$ is the vector of feature values of the instance to be explained.

Such calculated Shapley values show the contribution of the feature values of a particular instance to the prediction of the model with respect to the expected value.
This contribution can be positive or negative denoting if the particular value has a larger effect for the positive or negative class respectively, in the case of binary classification.
This information confronted with expert knowledge on how the values of different machinery sensors contribute to healthy and unhealthy conditions can help to determine if the classification model behaves as if it were a healthy state (also referring to the product shift) or as if it were a failure.

\begin{figure}[t]
\centering
\begin{minipage}{.5\textwidth}
  \centering
  \includegraphics[width=1\linewidth]{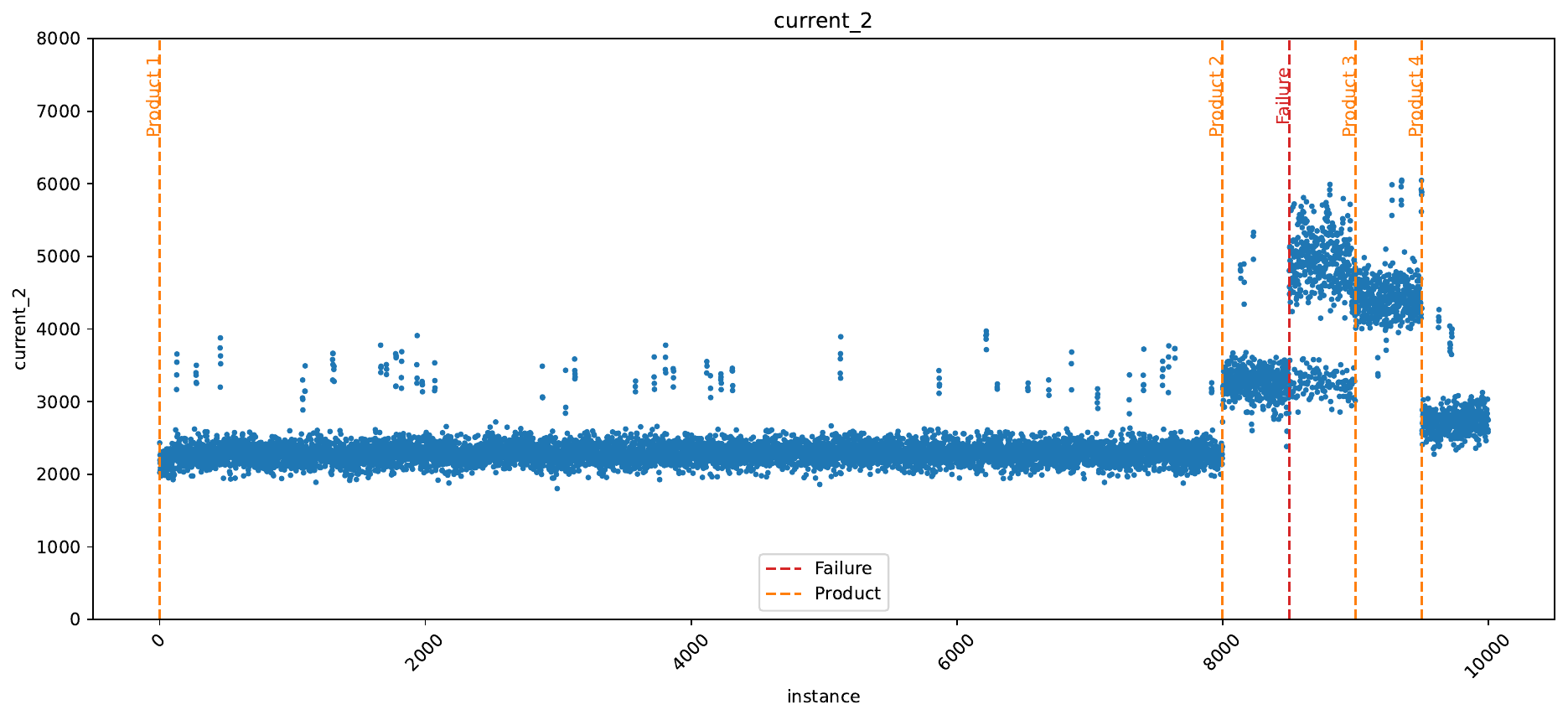}
\end{minipage}%
\begin{minipage}{.5\textwidth}
  \centering
  \includegraphics[width=1\linewidth]{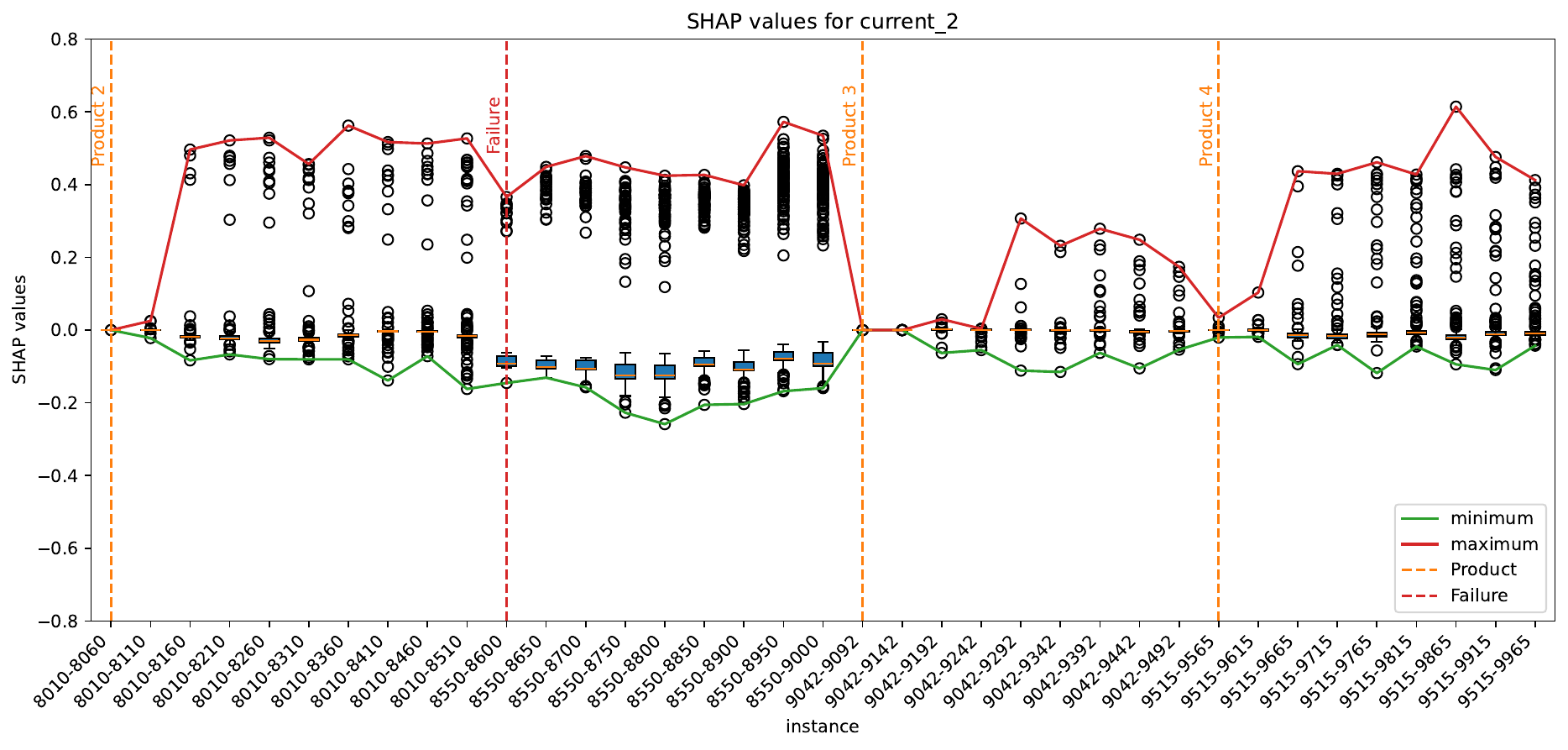}
\end{minipage}
\caption{Raw signal for \textit{current\_2} on the left, SHAP values plotted in batches for \textit{current\_2} on the right.}
\label{fig:example}
\end{figure}

To provide a broader spectrum of such a behavior of the model, we calculate and plot Shapley values over time and present them to the expert for a final decision on the nature of the change. 
The specialist should pay attention to the explanation of the features that are crucial for the proper operations of the production line. If there is an inconsistent with expert knowledge change in explanations for those key features, then the specialist should look at the raw signals and decide what is the root cause of the change. The example is presented in Figure~\ref{fig:example}. The expert indicates \textit{current\_2} as a key feature for anomaly detection in bearing degradation. When the failure begins, the median of SHAP values drops below zero, which should be an alert of unexpected bahavior. There is also a change of characteristics in raw signals for instances where a failure was identified. Adding the knowledge about e.g. previous repairs, the expert should decide if the change corresponds to the new product (domain shift) or failure. 

\section{Experiments}
\label{sec:exp}


This section presents a case study illustrating the use of our method in an industrial environment. 
We examined the cold rolling process in a steel plant, where differentiating between failures and new types of rolled products is crucial because of the variety of different products being rolled over time. 
The following subsections describe the cold rolling dataset and the experimental outcomes.
\subsection{Cold rolling mill dataset}
\begin{figure}[h]
\centering
\includegraphics[width=0.7\textwidth, scale=0.5]{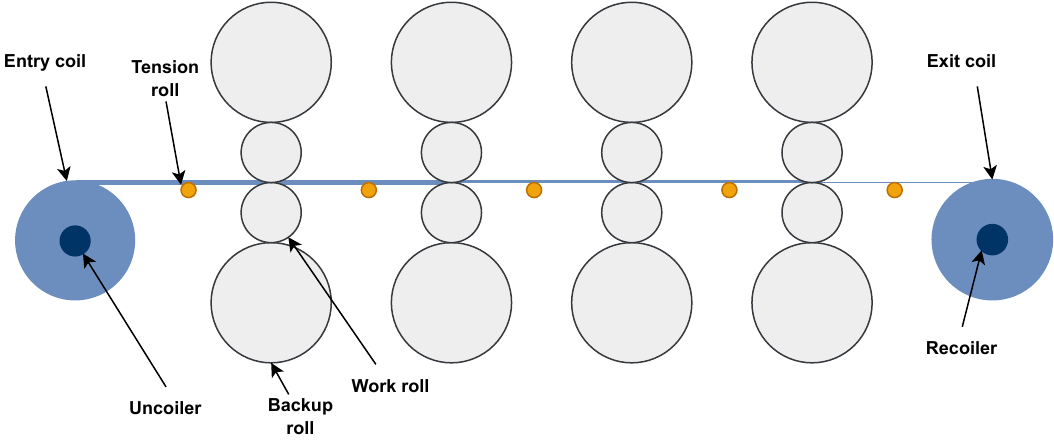}
\caption{The schematic diagram of four stand cold rollling mill.} \label{fig:CRM}
\end{figure}

The cold rolling process is responsible for reducing the strip's thickness by passing it through a pair of rolls. Initially, the steel coil is unrolled on an uncoiler. During rolling, the strip moves slowly between the rolling stands. Once the uncoiler grips the strip, the process speeds up to reach the desired thickness (Figure \ref{fig:CRM}). During the rolling process, various products different in mechanical properties, thickness, width and reduction are processed.
The analyzed dataset comes from a simulator of a four--stand rolling mill, which was created using real measurements from a steel company. The raw data samples are collected at regular time intervals, and all the signals are listed in Table \ref{tab:rolling_dataset}.
\begin{table}[h]
\centering
  \caption{Description of cold rolling mill dataset.}
  \label{tab:rolling_dataset}
   \scalebox{0.8}{
  \begin{tabular}{ccc}
    \toprule
     \textbf{Feature}&\textbf{Description}&\textbf{Unit}\\
    \midrule
    ithick/othick & coil thickness at the beginning/end of rolling &\unit{\milli\metre}\\
    width & coil width & \unit{\milli\metre}\\
    ys0/ys1 & yield strength at the beginning/end of rolling & \unit{\MPa}\\
    work\_roll\_diam & diameter of roll & \unit{\milli\metre}\\
    work\_roll\_mileage & mileage of roll & \unit{\milli\metre}\\
    reduction & reduction on specific stand & \%\\
    tension & tension on specific stand & \unit{\kN}\\
    roll\_speed & speed of roll on specific stand & \unit[per-mode = symbol]{\metre\per\second}\\
    force & force on specific stand & \unit{\kN}\\
    torque & torque on specific stand & \unit{N\metre}\\
    gap & gap between two rolls on specific stand & \unit{\milli\metre}\\
    current & current on specific stand & \unit{\ampere}\\
  \bottomrule
\end{tabular}
}
\end{table}
For experimental purposes, 10000 samples were acquired containing parameters from 4 types of rolled products. The dataset composition was chosen to reflect the real--world situation, where most of the production consists of one type of product (or related). Each product in the dataset is described by three parameters: \textit{ithick/othick/width}. The anomalies represent abnormalities in the mechanical degradation of bearings. The expert from this particular rolling infrastructure indicates electric \textit{current} and mechanical \textit{torque} as the most important parameters responsible for bearings degradation. The dataset composition is presented below:
\begin{itemize}
\item \textbf{Product 1} --\textit{3.4/1.61/918.67}, the major (source) product which constitutes 80\% of the dataset, about 8.5\% anomalies corresponding to bearings on all stands.
\item \textbf{Product 2} --\textit{3.0/1.11/1082.43}, constitutes 5\% of the dataset, about 10.8\% anomalies corresponding to bearings on all stands.
\item \textbf{Failure on Product 2} -- \textit{3.0/1.11/1082.43}, constitutes 5\% of the dataset, about 77.4\% anomalies simulating bearings failure on stand 2.
\item \textbf{Product 3} --\textit{2.8/0.82/918.58}, constitutes 5\% of the dataset, about 6\% anomalies corresponding to bearings on all stands.
\item \textbf{Product 4} --\textit{3.5/1.44/1080.20}, constitutes 5\% of the dataset, about 12.6\% anomalies corresponding to bearings on all stands.
\end{itemize}

These elements appear in this order in the analyzed stream. Note that products 2,3 and 4 are rather rare events compared to the much longer production of the main product 1. This is similar to research on class imbalanced and concept drifting streams, which are particularly difficult for most learning algorithms, see \cite{brzezinski2021impact}.

\subsection{Classical approaches for anomaly detection}

\begin{figure}[t]
\centering
\begin{minipage}{.5\textwidth}
  \centering
  \includegraphics[width=1\linewidth]{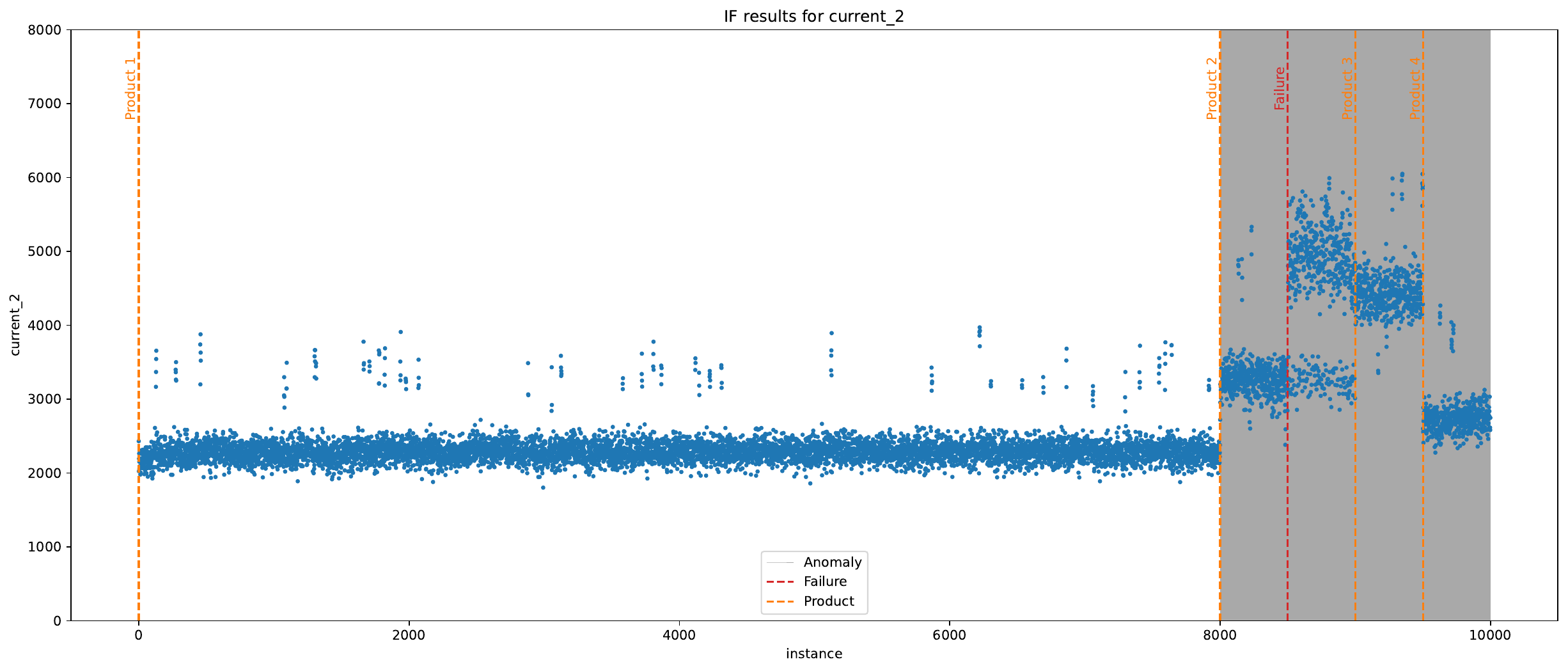}
\end{minipage}%
\begin{minipage}{.5\textwidth}
  \centering
  \includegraphics[width=1\linewidth]{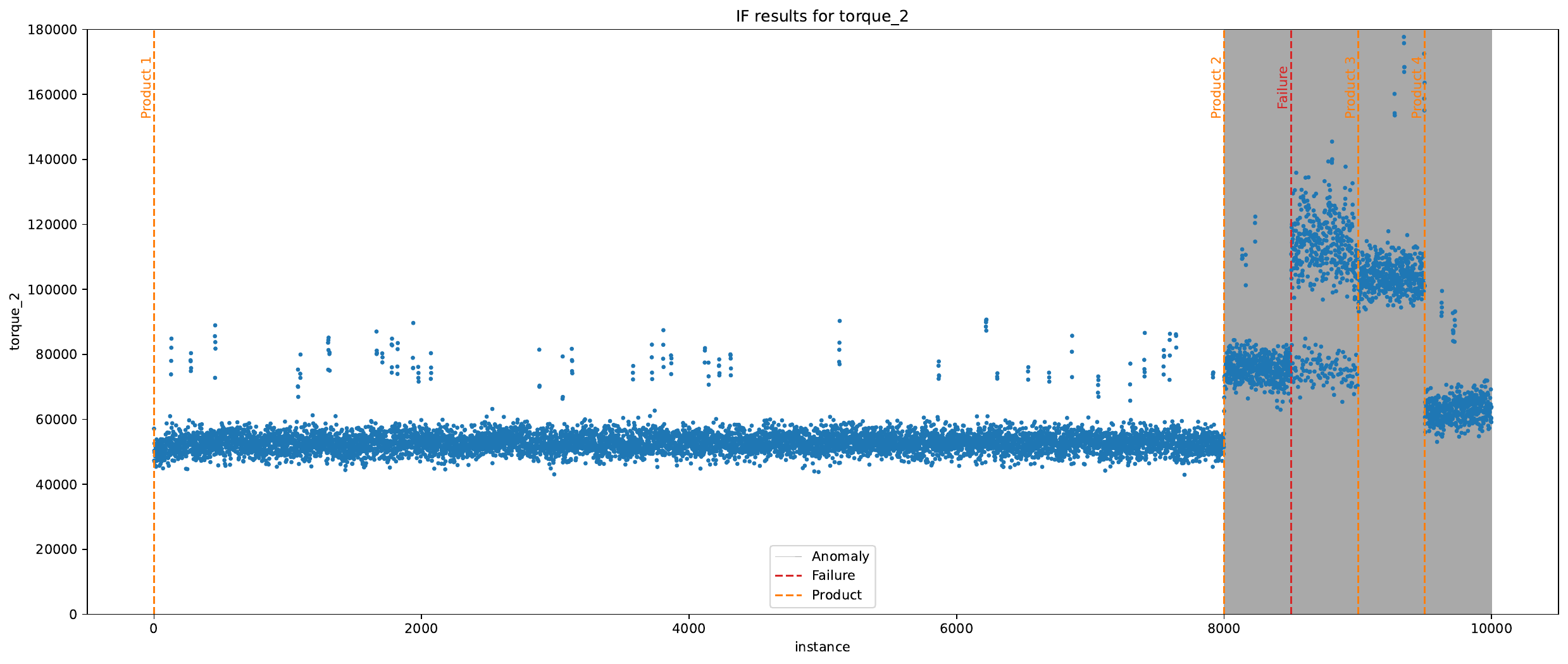}
\end{minipage}

\vspace{0.3cm}

\centering
\begin{minipage}{.5\textwidth}
  \centering
  \includegraphics[width=1\linewidth]{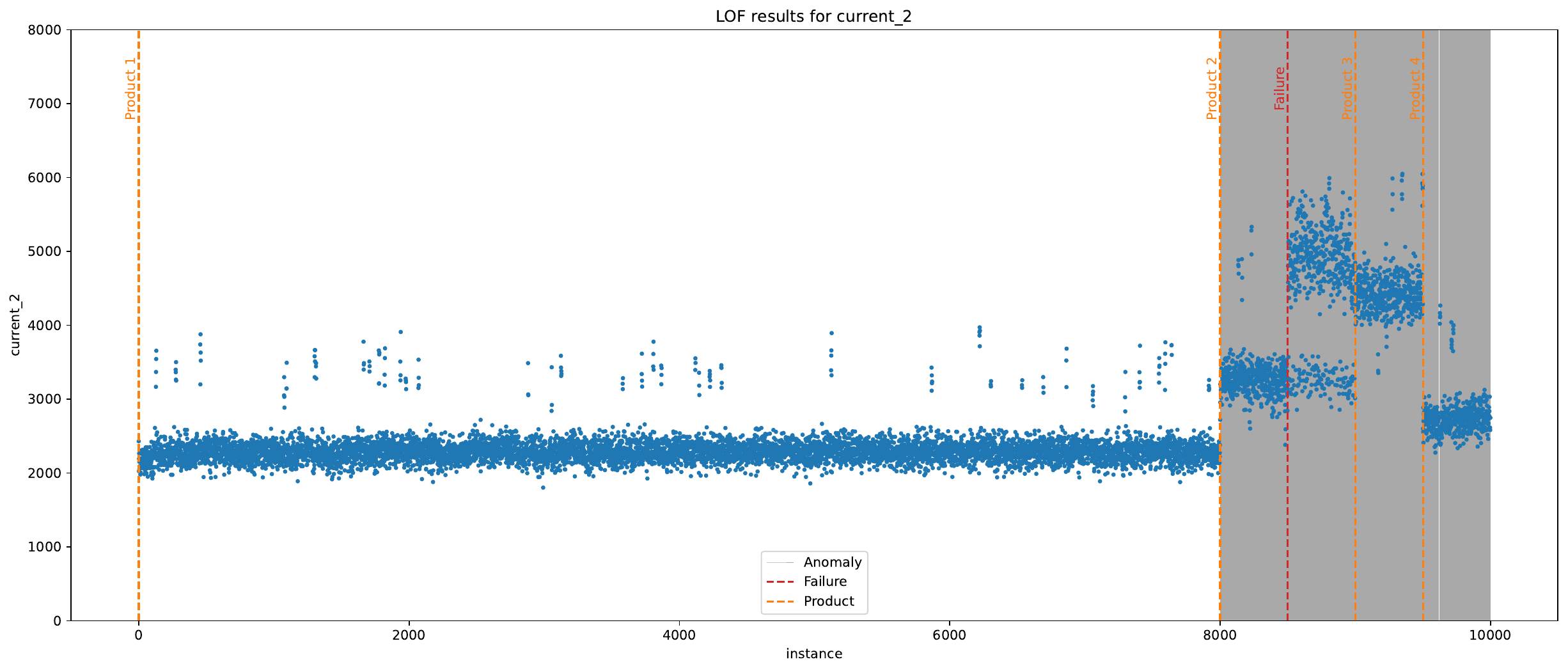}
\end{minipage}%
\begin{minipage}{.5\textwidth}
  \centering
  \includegraphics[width=1\linewidth]{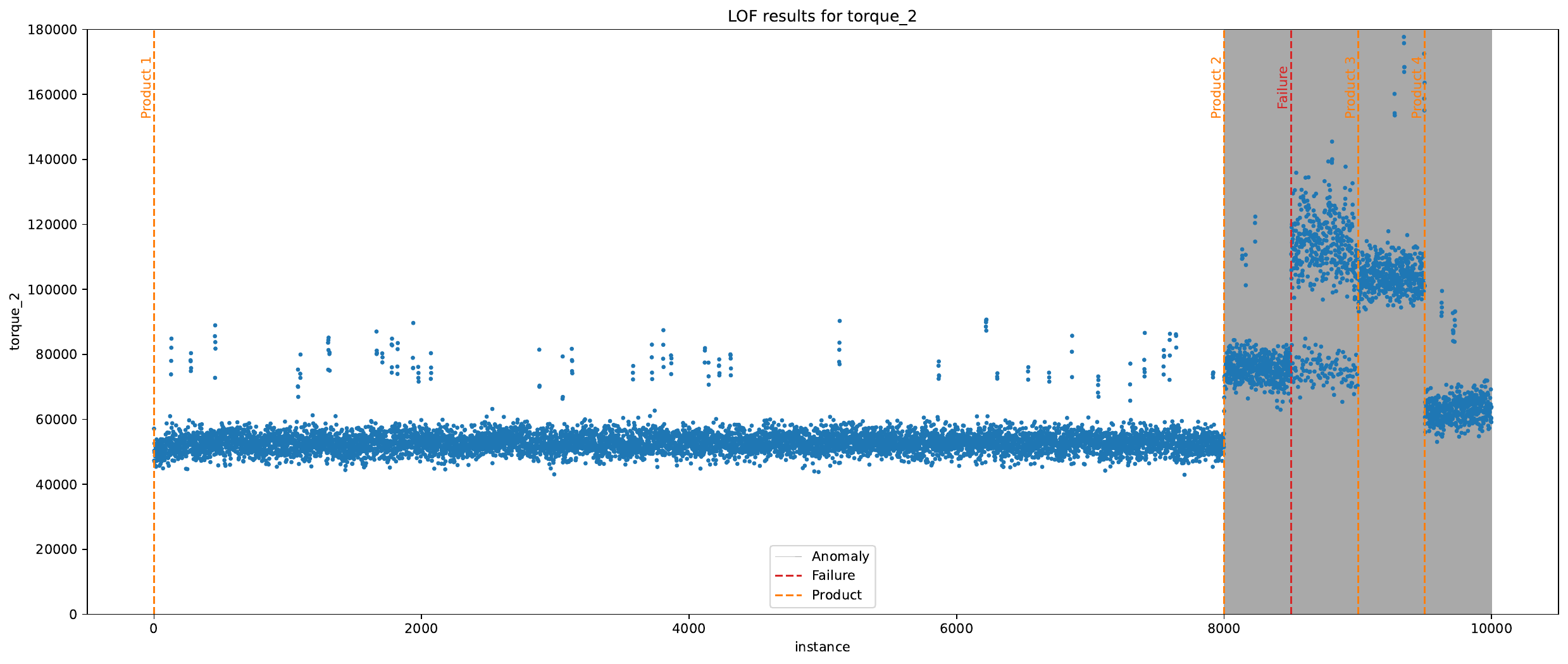}
\end{minipage}%

\vspace{0.3cm}

\centering
\begin{minipage}{.5\textwidth}
  \centering
  \includegraphics[width=1\linewidth]{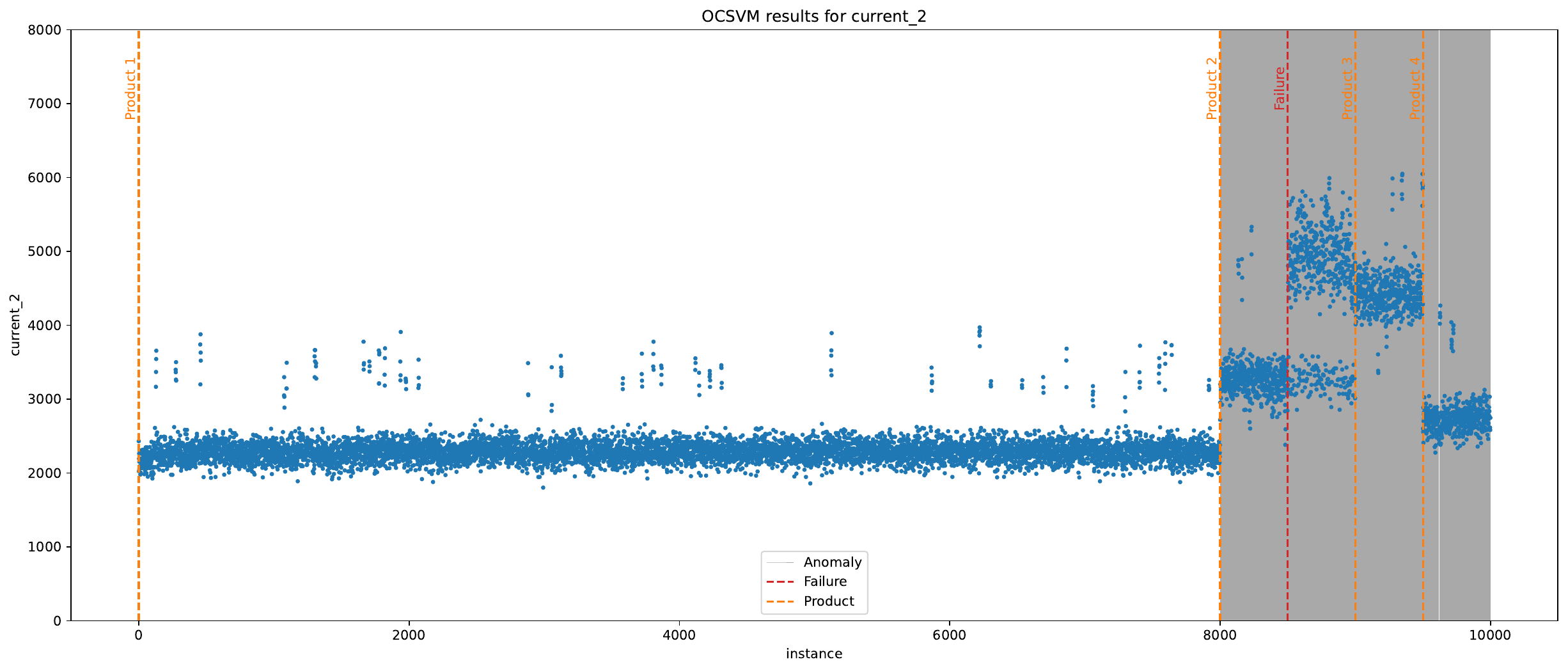}
\end{minipage}%
\begin{minipage}{.5\textwidth}
  \centering
  \includegraphics[width=1\linewidth]{figures/OCSVM_current_2.pdf}
\end{minipage}%

\vspace{0.3cm}

\centering
\begin{minipage}{.5\textwidth}
  \centering
  \includegraphics[width=1\linewidth]{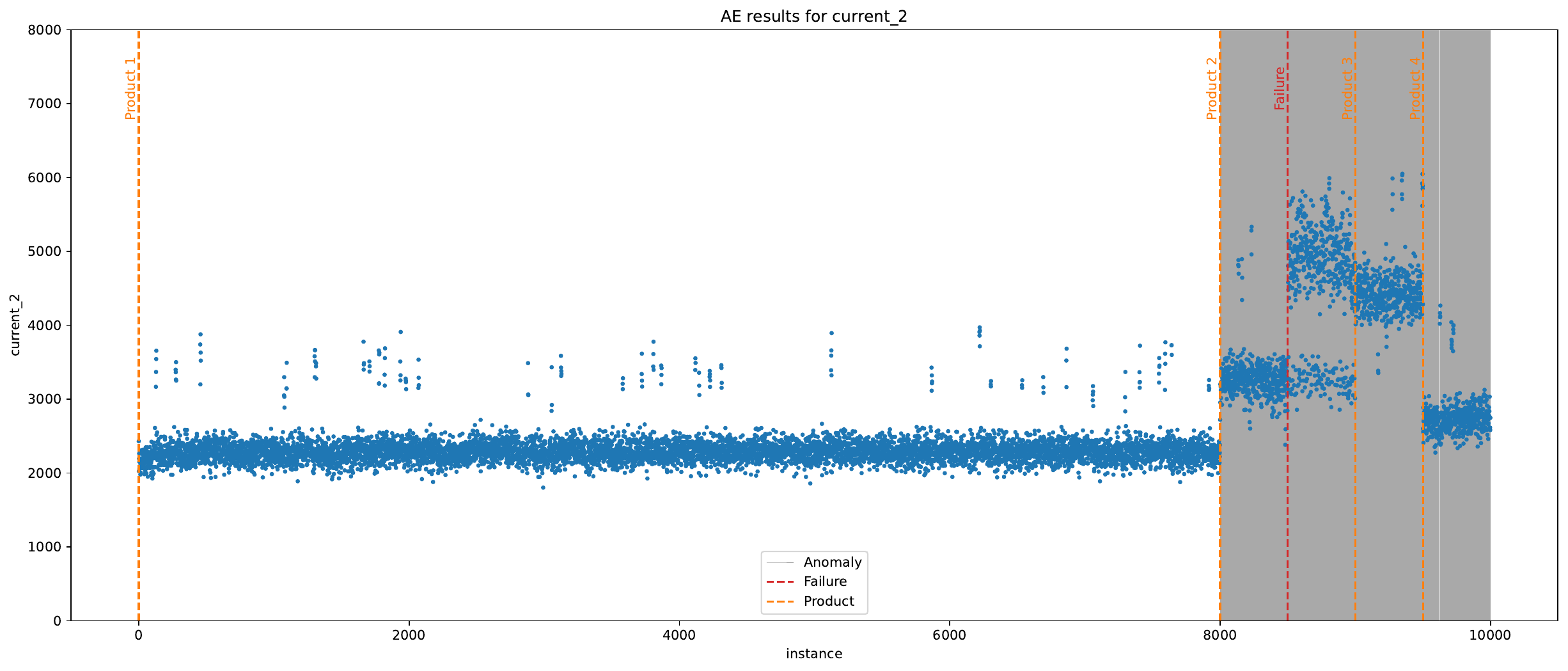}
\end{minipage}%
\begin{minipage}{.5\textwidth}
  \centering
  \includegraphics[width=1\linewidth]{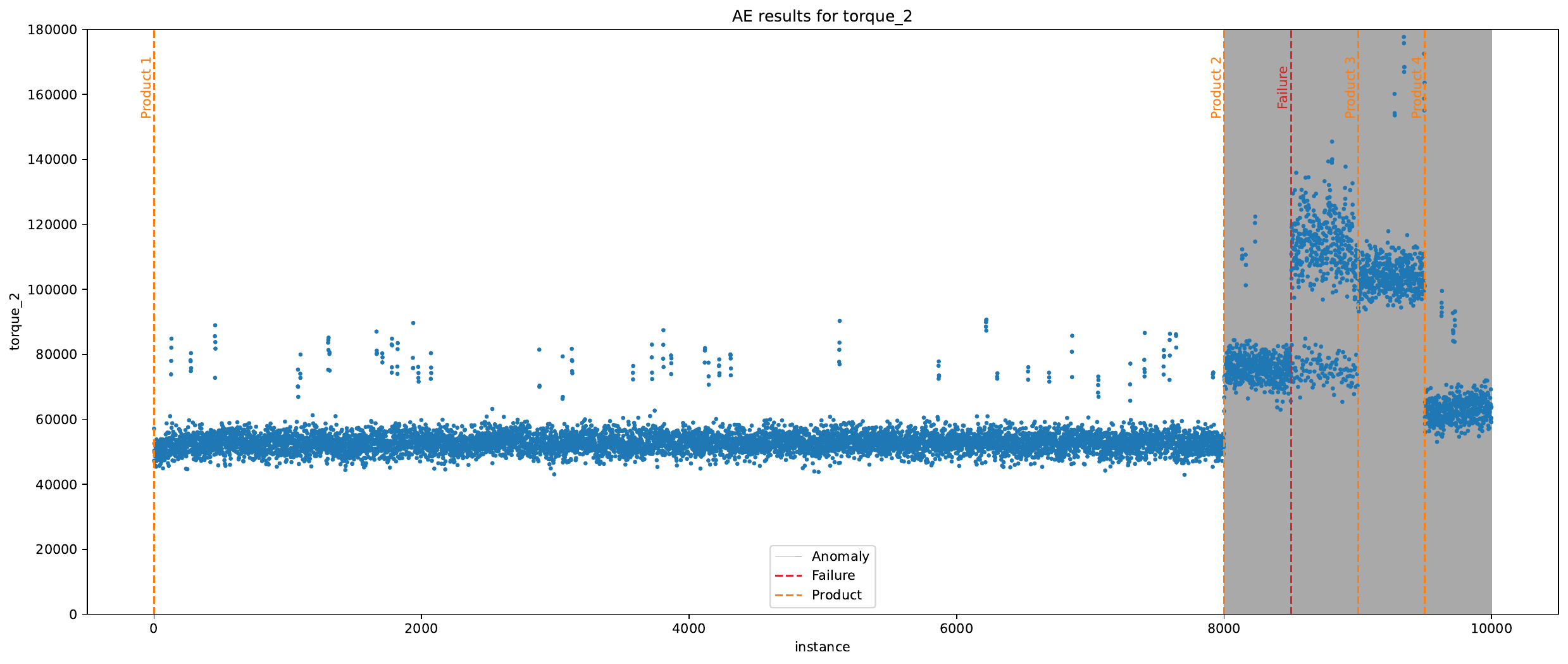}
\end{minipage}%
\caption{Anomaly detection algorithms evaluation on target products on important signals from rolling stand 2.}
\label{fig:anomaly_results}
\end{figure}

For detecting anomalies we chose the state-of-the-art techniques like Isolation Forest (IF), Local Outlier Factor (LOF) and One Class Support Vector Machine (OCSVM), Autoencoder (AE). The models were trained on signals from the source product (Product 1) and evaluated on the rest of the stream (target products). In the Figure~\ref{fig:anomaly_results} the algorithms results on \textit{current} and \textit{torque} from stand 2 is presented. The dashed lines denote instances when a new product/failure appears on the production line. Grey lines signify instances where anomalies were detected. It is observable that all tested algorithms tend to classify both products and failures as anomalies. The potential reason is that the measurement characteristic is different from the source product for both of those cases. Consequently, it becomes exceedingly challenging for anomaly detection algorithms to discriminate between failures and newly introduced rare products. 

\subsection{The proposed method}
In the subsequent section, we present experiments conducted using the proposed method, which leverages transfer learning and explainable artificial intelligence. 

\begin{figure}[h]
\centering
\includegraphics[scale=0.28]{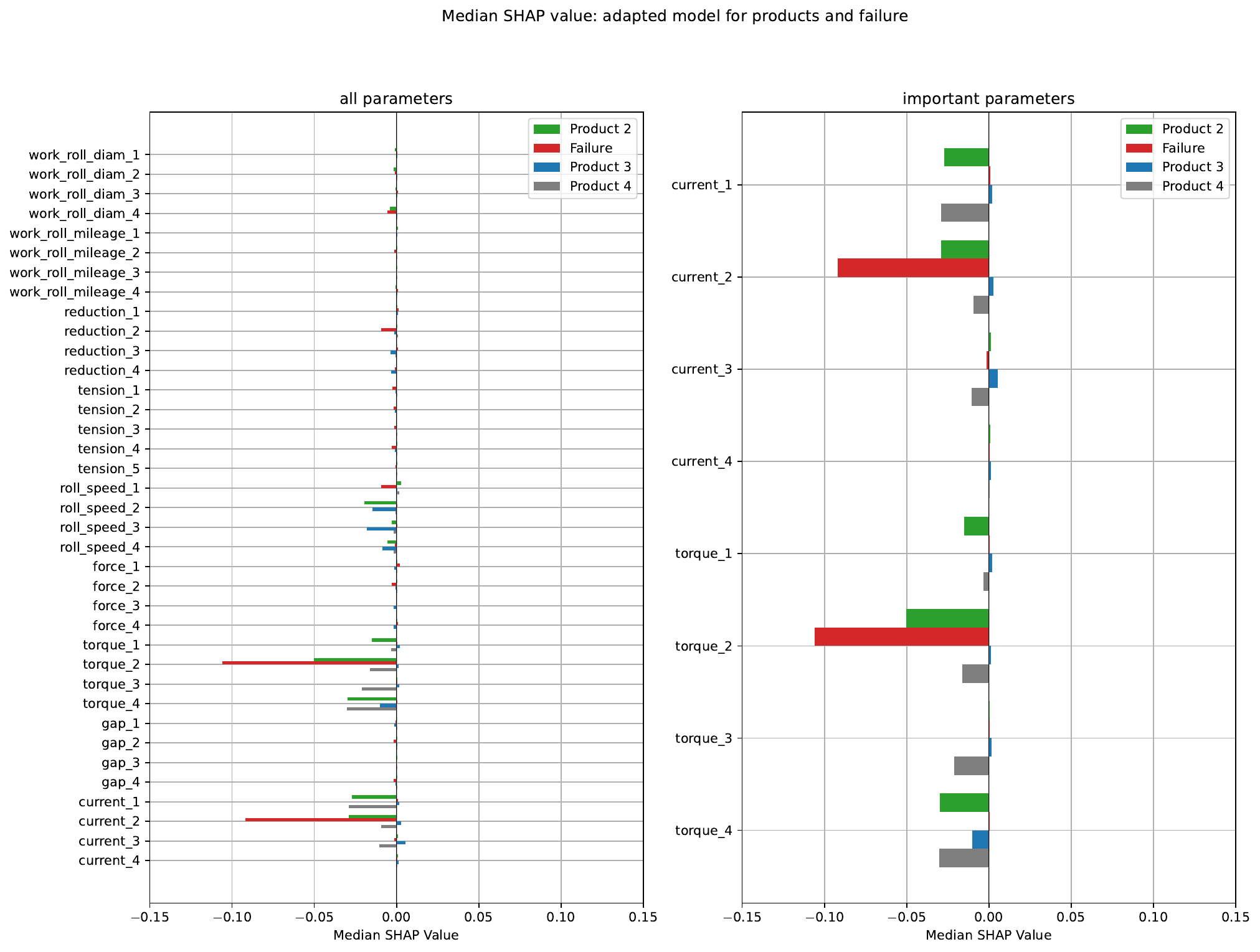}
\caption{SHAP values median for: a) all parameters (on left), b) for important parameters for anomaly detection (on right).} 
\label{fig:shap_median}
\end{figure}

Firstly the domain adaptation from the source product to each target is conducted. For domain adaptation, we selected the feature--based classification and contrastive semantic alignment (CCSA) algorithm~\cite{motiian2017unified}. The CCSA goal is to create an encoded space where the distances between source and target pairs with the same label are minimized, while the distances between pairs with different labels are maximized. This model was chosen because semantic alignment algorithms achieve good results even with a small number of samples in the target domain. Subsequently, the SHAP explainer is applied to each target domain, and the median SHAP values for each parameter are computed. This approach aims to determine if there are differences in SHAP explanations (i.e. median feature importance) between products and failures. In the Figure~\ref{fig:shap_median} can be observed that adaptation for failure leads to clearly different median SHAP values for the parameters \textit{current\_2} and \textit{torque\_2} which expert has identified as the most critical for failure diagnosis.

\begin{figure}[h]
\centering
\includegraphics[scale=0.27]{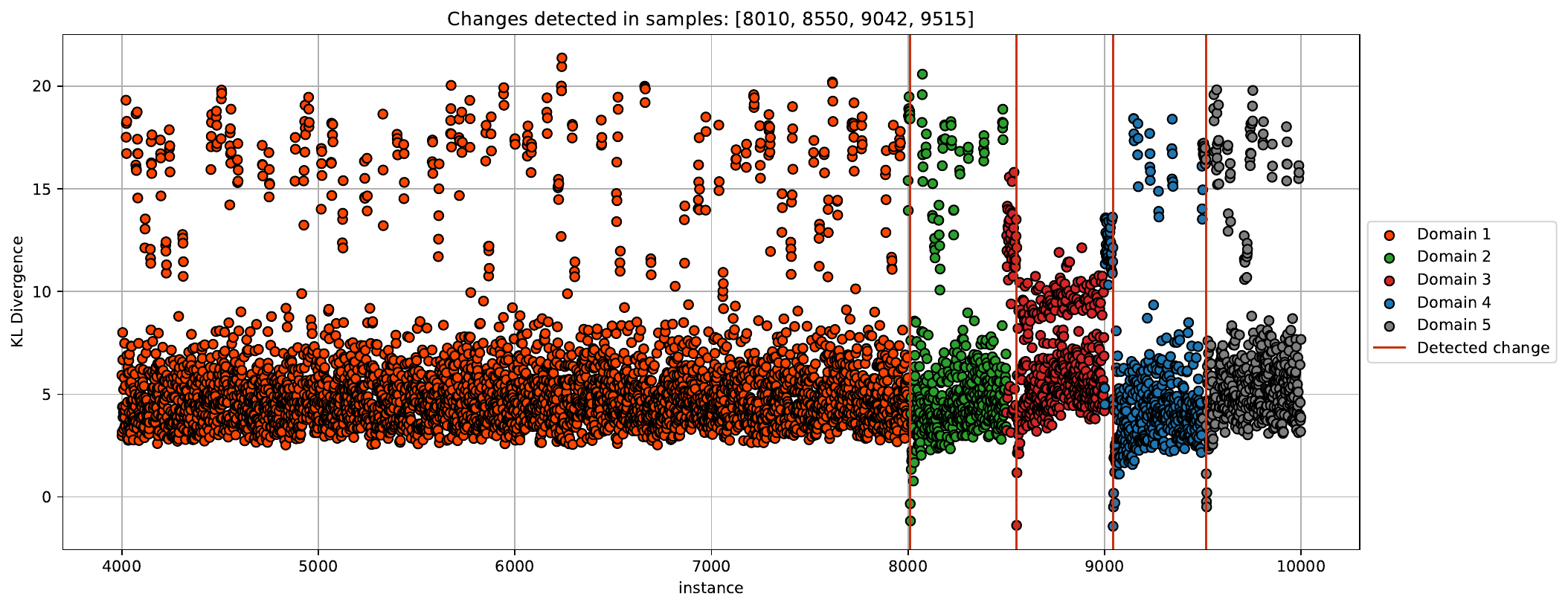}
\caption{Detected changepoints in the dataset.} 
\label{fig:KL}
\end{figure}

\begin{figure}[h]

\centering
\begin{minipage}{.5\textwidth}
  \centering
  \includegraphics[scale=0.2]{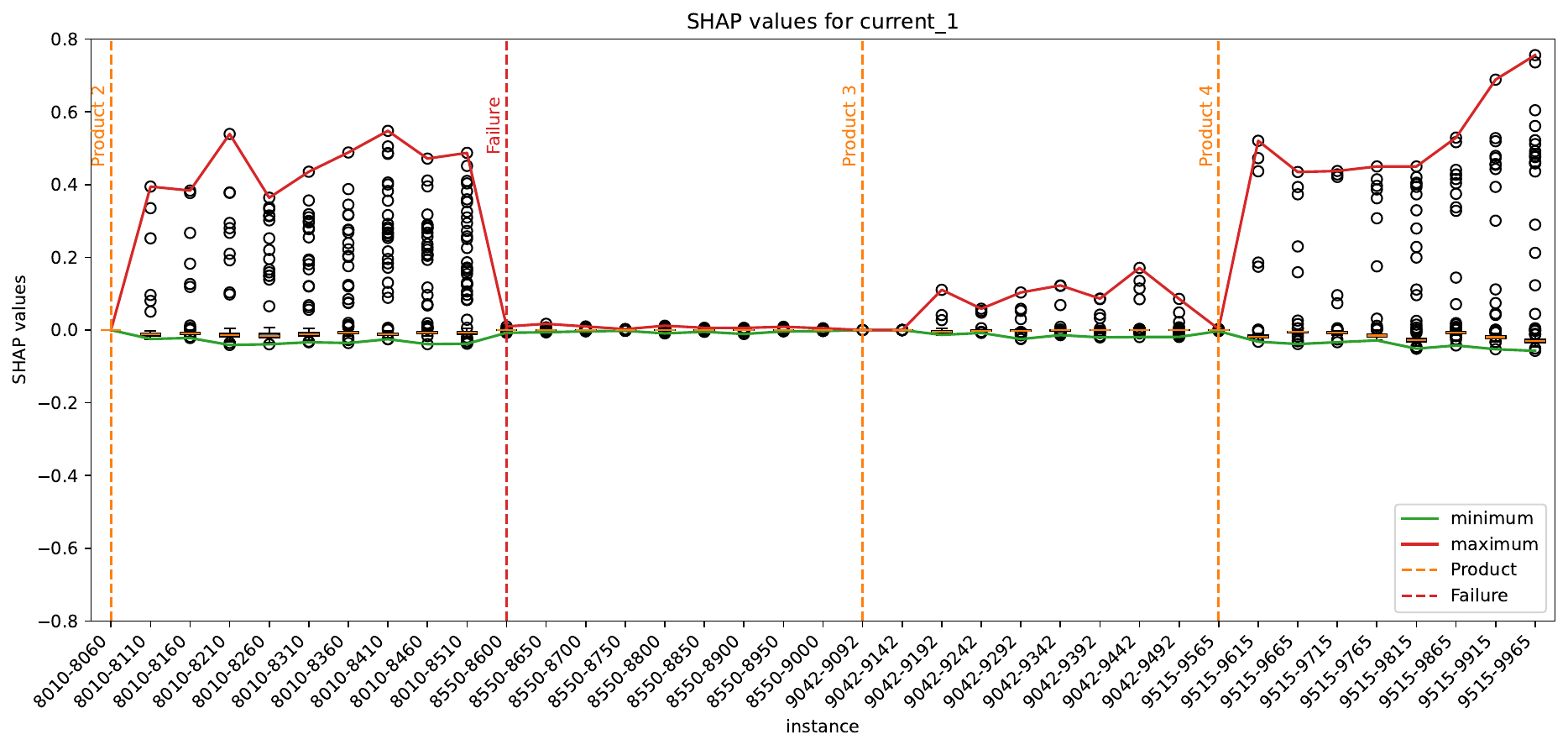}
\end{minipage}%
\begin{minipage}{.5\textwidth}
  \centering
  \includegraphics[scale=0.2]{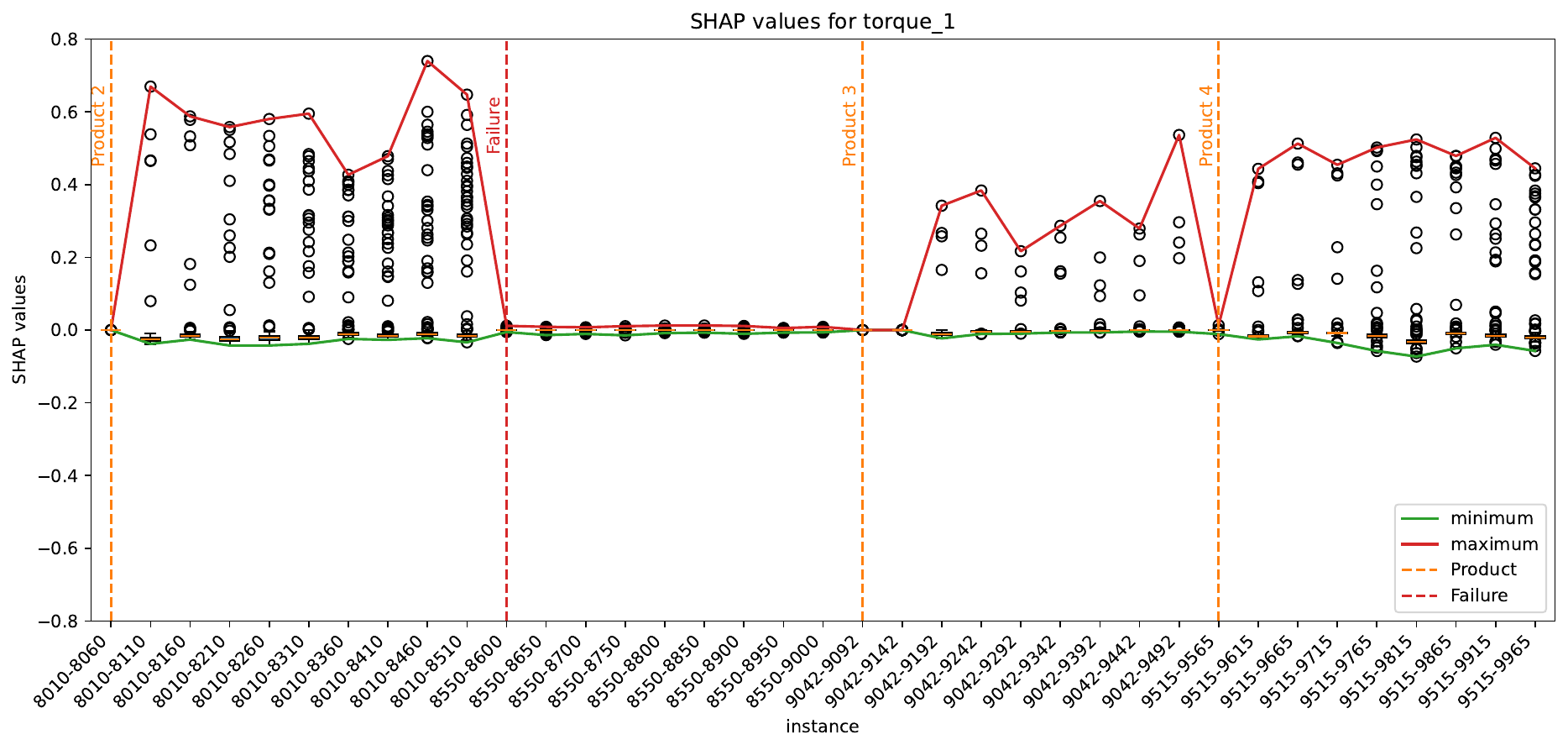}
\end{minipage}

\vspace{0.3cm}

\centering
\begin{minipage}{.5\textwidth}
  \centering
  \includegraphics[scale=0.2]{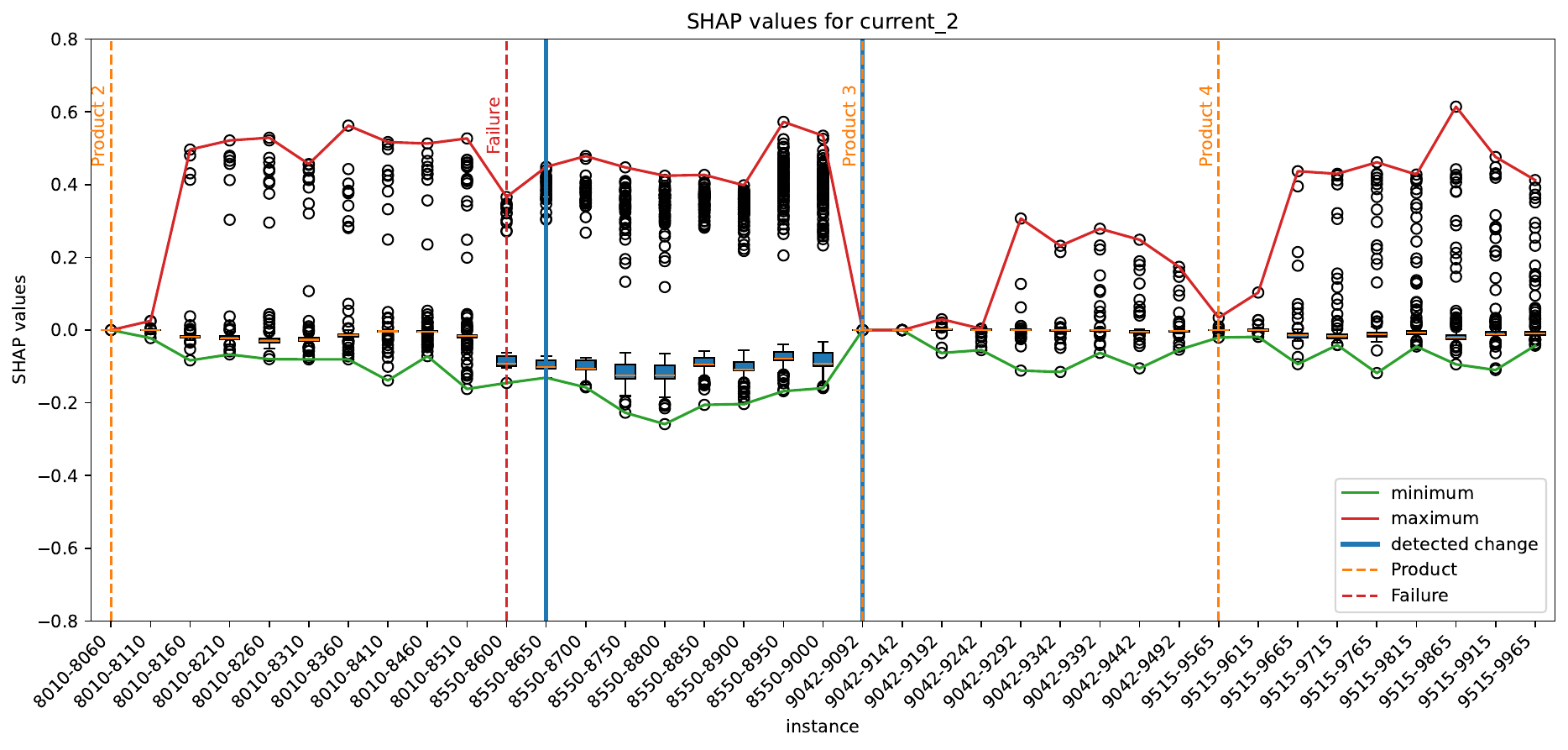}
\end{minipage}%
\begin{minipage}{.5\textwidth}
  \centering
  \includegraphics[scale=0.2]{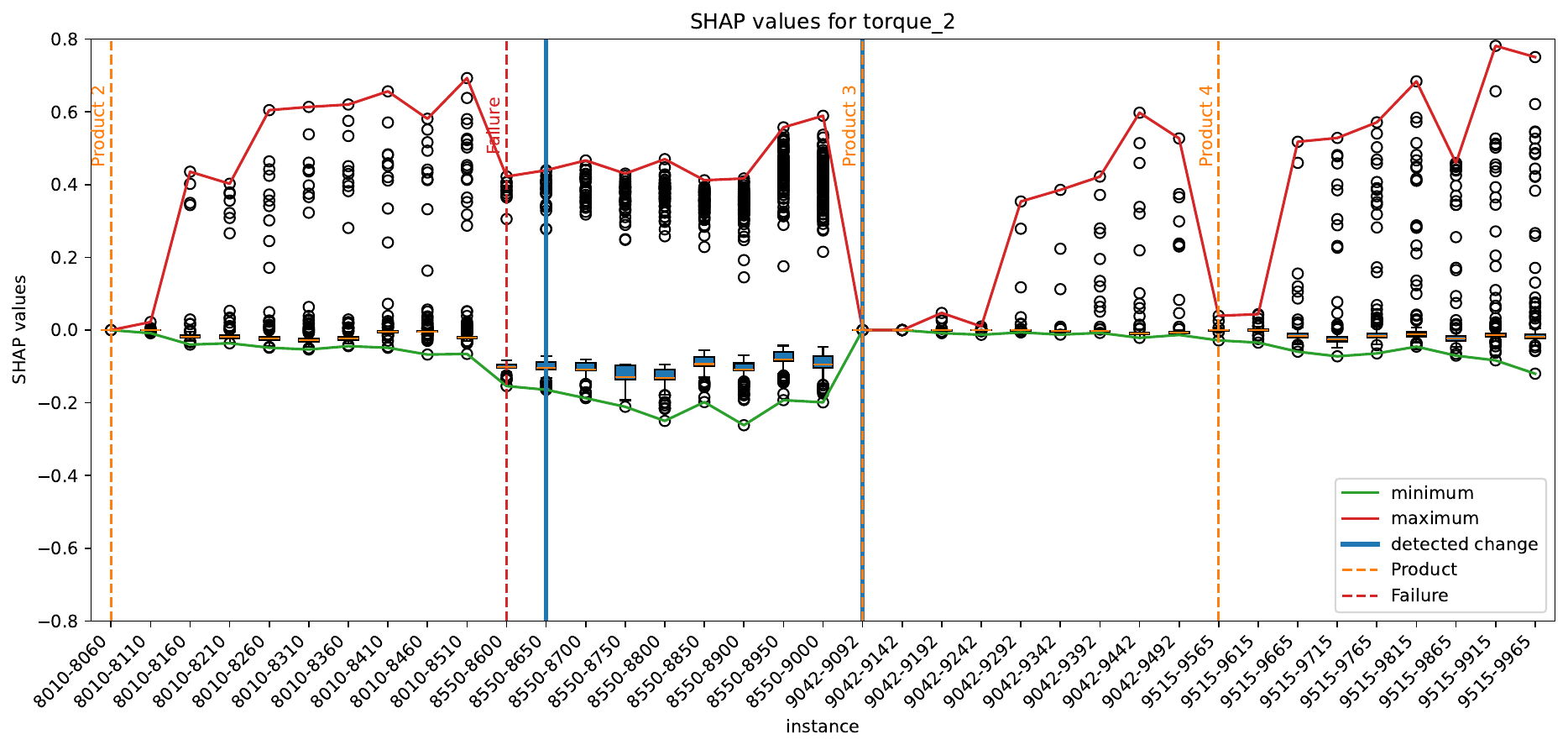}
\end{minipage}%

\vspace{0.3cm}

\centering
\begin{minipage}{.5\textwidth}
  \centering
  \includegraphics[scale=0.2]{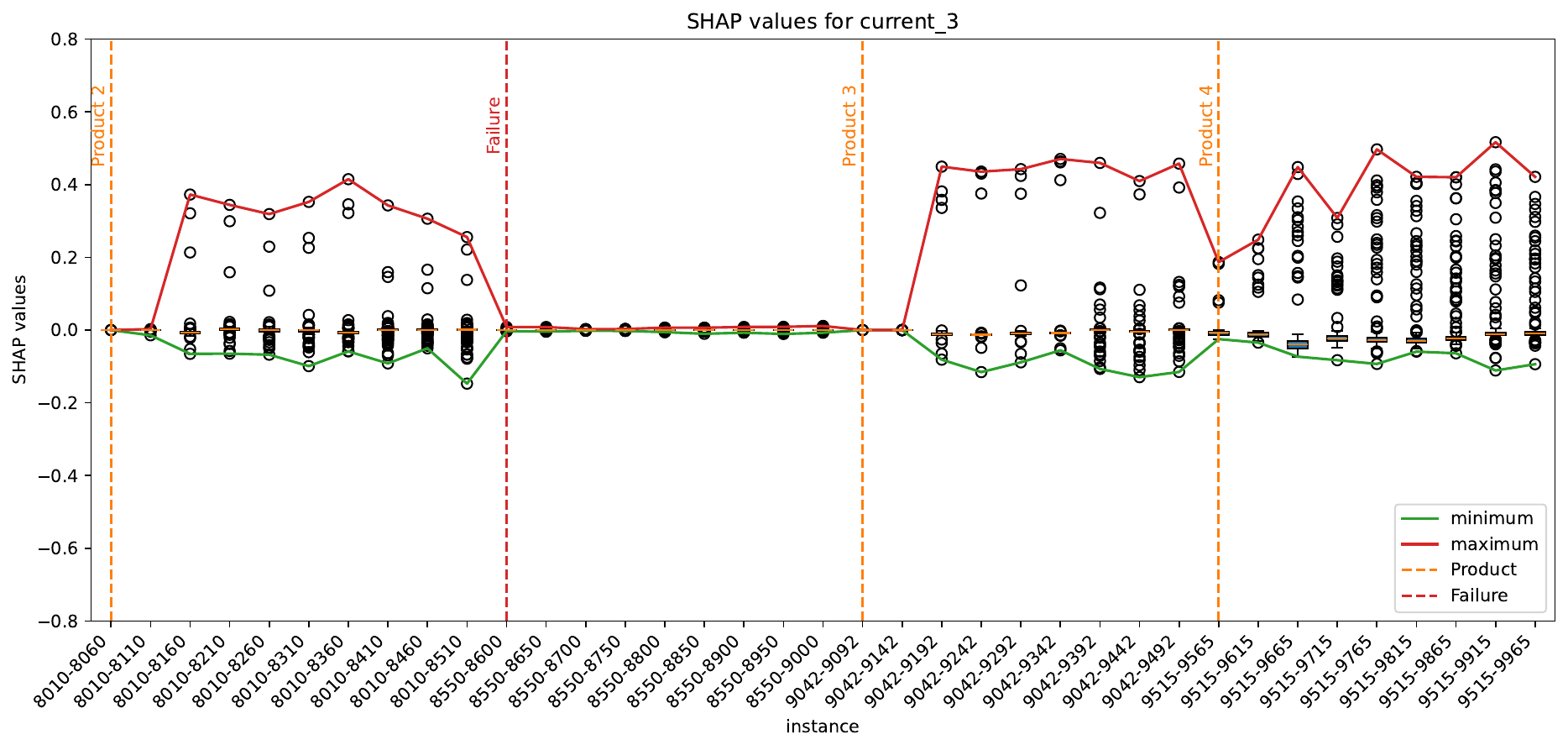}
\end{minipage}%
\begin{minipage}{.5\textwidth}
  \centering
  \includegraphics[scale=0.2]{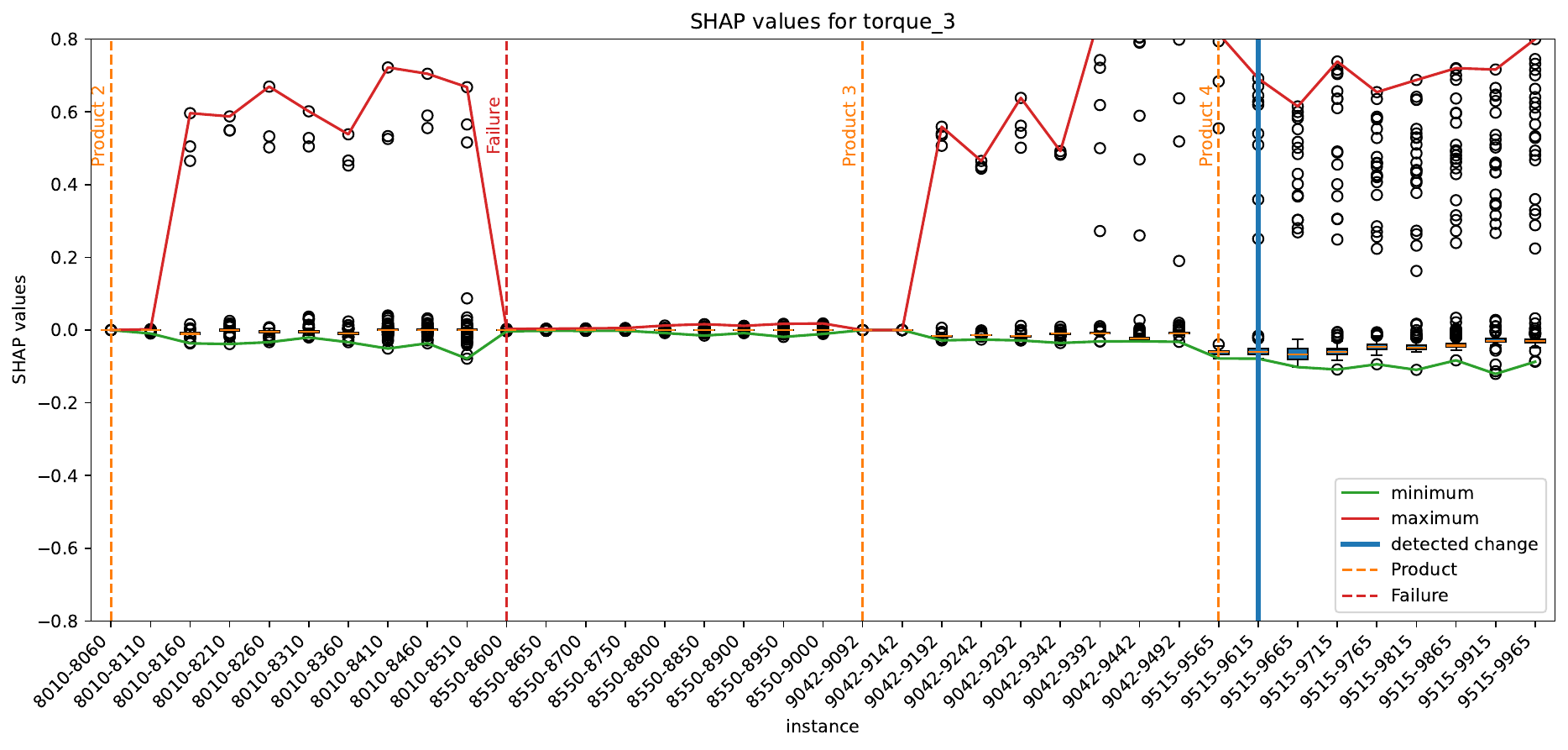}
\end{minipage}%

\vspace{0.3cm}

\centering
\begin{minipage}{.5\textwidth}
  \centering
  \includegraphics[scale=0.2]{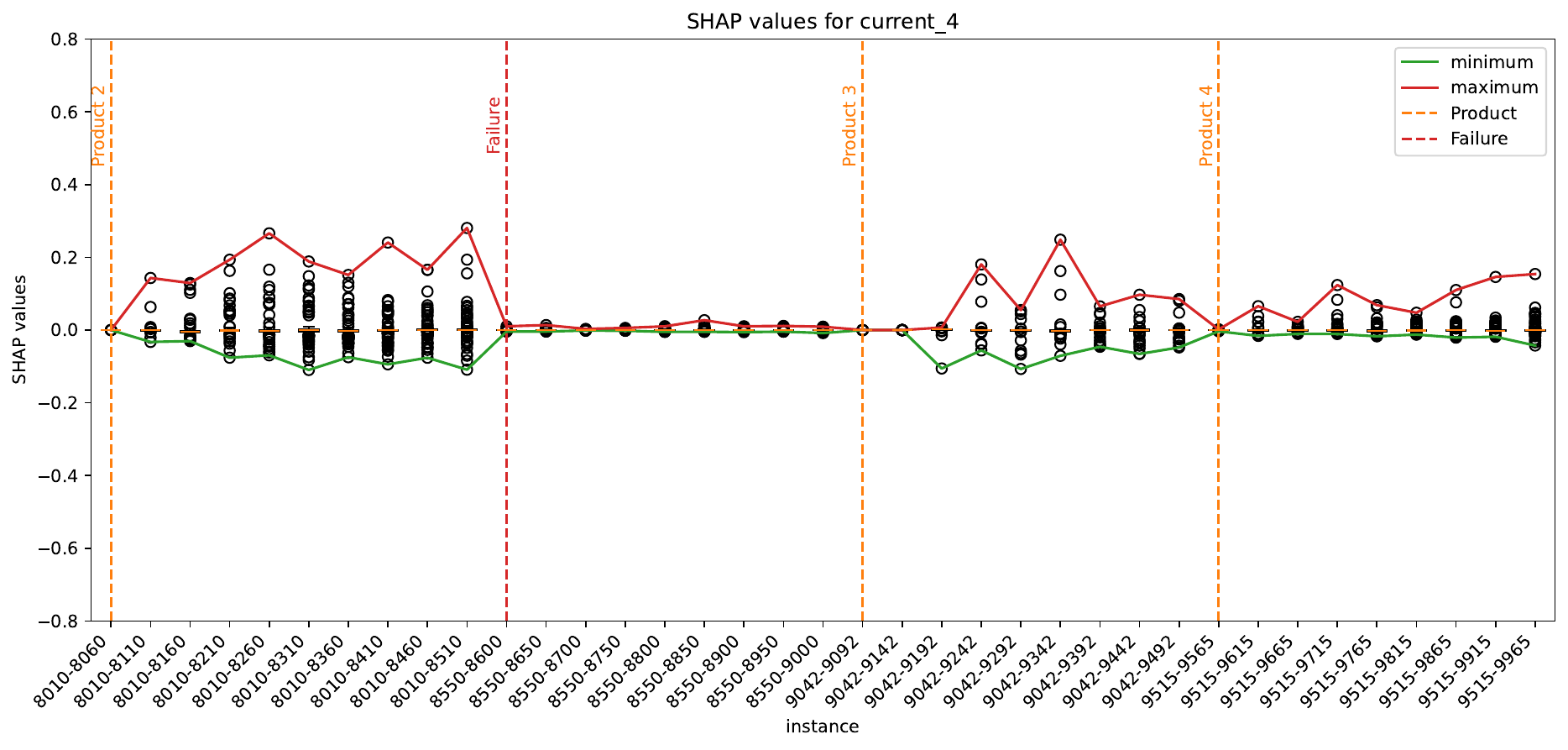}
\end{minipage}%
\begin{minipage}{.5\textwidth}
  \centering
  \includegraphics[scale=0.2]{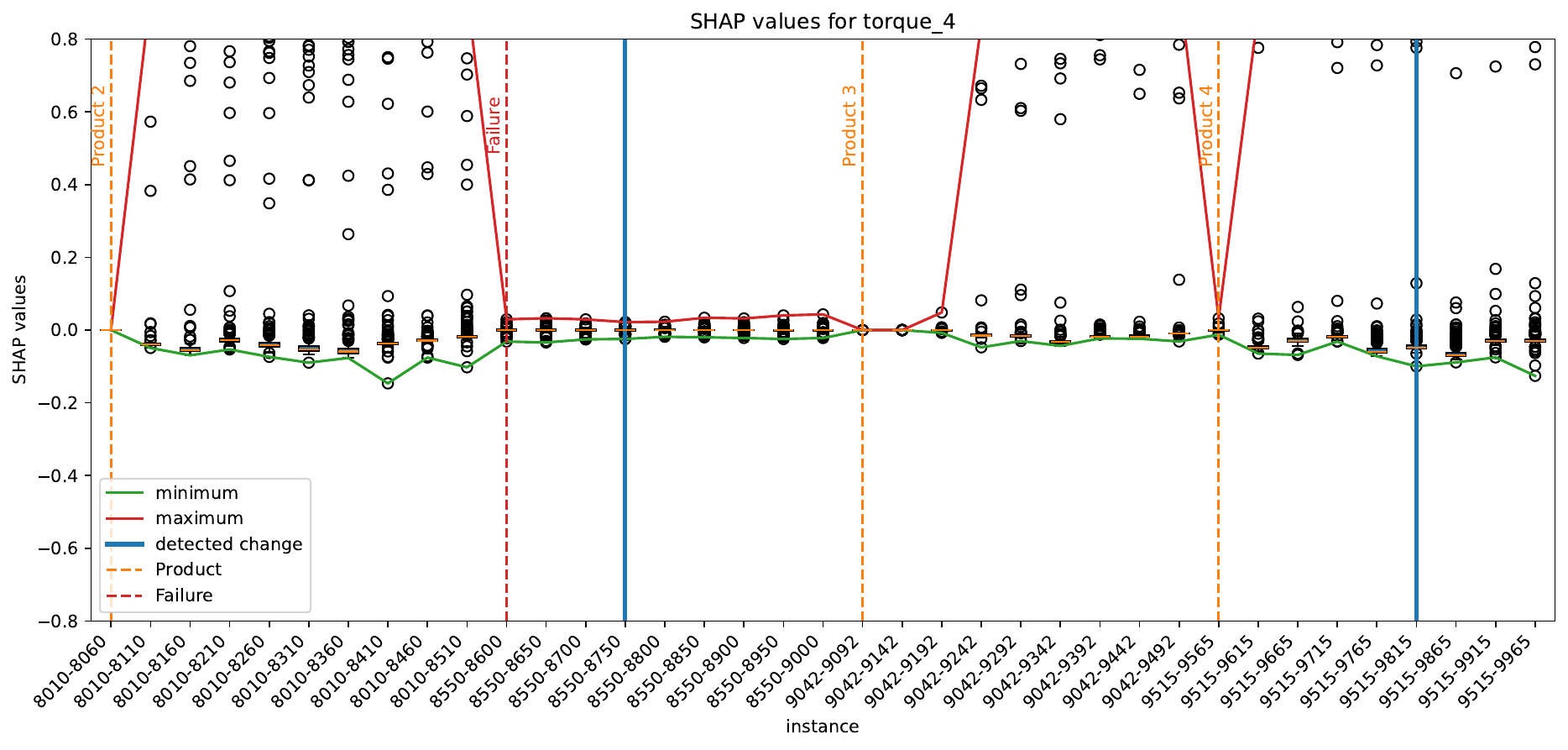}
\end{minipage}%
\caption{Boxplots of SHAP values important for bearing diagnosis.}
\label{fig:boxplot_in_stream}
\end{figure}

The next step was to explore whether the explanations of specific parameters changed during the rolling process and in what places it occurred. The experiment started by calculating the KL divergence using the formula \ref{eq:KL} and applying PH drift detector. The experimental parameters were defined as follows: 40\% of the dataset was designated as the reference distribution, and the features used for calculation included the \textit{current} and \textit{torque} on each stand. Such defined changepoint detector marked changes in samples 8010, 8550, 9042, and 9515 when the real points were in 8000, 8500, 9000 and 9500. The outcome of applying the PH drift detector to the estimated KL can be observed in Figure~\ref{fig:KL}. 
Upon detecting changepoints, we adapted from source to target products in batches. Specifically, the CCSA model was retrained after each batch of incoming samples from the target domain (demonstrated with batch\_size=50 below). Each time a new changepoint was detected, the adaptation process was restarted.
After model training, a SHAP explainer was constructed using the training batch of samples. Subsequently, the explanation of specific parameters was presented as a box plot for each training batch to monitor the evolution of SHAP values (Figure~\ref{fig:boxplot_in_stream}).  Furthermore, the maximum (red line) and minimum (green line) SHAP values were visualised. The plots clearly indicate that for \textit{current\_2} and \textit{torque\_2}, there is a shift in SHAP value explanations as the model begins to adapt to failure conditions. The median and minimum SHAP values decrease compared to those observed during the adaptation to products. Moreover, it is notable that the SHAP explanations for other rolling stands exhibit reduced variability during failure. To assess whether changes could be detected by an automated method, the Page-Hinkley (PH) drift detector was applied to the medians of the SHAP values. While drifts were identified in stand 2, additional detected changes in \textit{torque\_3} and \textit{torque\_4} were observed, which are likely false positives resulting from the sensitivity of the PH detector. 

The proposed method is designed to assist experts in detecting potential failures when raw data points are insufficient. In cases of some products, the signal may closely resemble the failure state, making it challenging to definitively classify by operator as healthy or failure. The evolving boxplots significantly change their characteristic when the domain adaptation model begins adapting to failure conditions. The abrupt decrease or increase in the median of SHAP values should alert experts and be a starting point for in--depth analysis. By examining these plots, experts in specific rolling infrastructures could combine our analysis with raw signals and the machinery condition (e.g. recent bearing replacement, last renovation etc.) to determine if a failure is beginning in the rolling mill. Automating the decision--making process to distinguish between failures and new products remains a direction for future research.

\section{Discussion and summary}
\label{sec:summary}
In this work, we propose a method that allows for human-guided differentiation between failures and healthy domain shifts.
We presented the feasibility of our approach on a dataset from a cold rolling facility of a steel factory and published the source code along with the dataset used in experiments for full reproducibility.
We demonstrated that with state-of-the-art anomaly detection methods, it is not possible to distinguish between healthy domain shift and failure.
On the other hand, our approach adds an additional information layer that human operators can use to make a more informed decision about the nature of changes in the data.
We consulted the results with an expert from the steel factory, who confirmed the correctness of the results obtained with our method.

Although our primary focus in this work was on the industrial application of the method, it is more generic and can be extended to other cases where the following conditions are met:
(1) The data come sequentially and may represent different distributions (domain shifts). 
(2) There are some hidden similarities between the distributions that can be captured by domain adaptation (transfer learning) algorithms.
(3) The data is analysed in batches, and in each batch, the majority of samples are considered candidates for the healthy / failure distribution, while the minority is considered anomalous.
These conditions can also be found in other areas where the data exhibit dynamic and non-stationary characteristics and adaptability is one of the features of the system.
In particular, such a situation is present in various predictive maintenance applications where typical maintenance actions such as replacements of an old component may cause changes in the data that result in the degradation of existing AI models~\cite{jakubowski2024artificial}.
Another example is network traffic monitoring~\cite{pietraszek2004intrusion} or electricity grid monitoring~\cite{gama2004drift}, where many different changes in the data appear that may reflect the normal change in electricity or bandwidth demand or a failure or attack. 
Reaching beyond the industrial cases in patient health monitoring, one can observe the same challenges, especially in online, long-term monitoring of patient health state, smart home installations, or ambient assisted living setups, where changes in habits may be confused with life-risk situations~\cite{falldetection2023}.

Future works can include many different path directions, as outlined in the following paragraphs.
In the current version of the method, the human operator is responsible for the analysis of the explanations.
This process can be automated to some extent by the analysis of the dynamics of explanations over time, instead of raw feature values.
This technique has been shown to be effective in similar use cases, as described in~\cite{kuk2023shaprootcause}.

The other direction is to focus on concept drift instead of domain shift.
Recent works approach concept drift handling with the usage of domain adaptation techniques~\cite{karimian2024dadrift}.
However, they mostly address the problem of performance of the model, not touching the explainability aspect.
Extending the work presented in our research with concept drift detection and explanation can scale the applicability of the approach to areas where keeping human in the loop is crucial (e.g. medicine, healthcare, etc.).

Finally, one of the limitations of the presented approach, which is also one of the hardest problems to handle in streaming data, is proper changepoint identification in the case of gradual data shifts.
Furthermore, the work presented in this aims primarily to distinguish between failures and domain shifts.
It does not explicitly explain the root causes of the failures or the domain shifts.
In this work, we discuss the preliminary results of our method, and in future works, we aim to address all of the above.






\begin{acknowledgments}
This paper is part of a project that has received funding from the European Union's Horizon Europe Research and Innovation Programme, under Grant Agreement number 101120406. The paper reflects only the authors' view and the EC is not responsible for any use that may be made of the information it contains. 

Jerzy Stefanowski's work was supported by the National Science Centre (Poland) grant No. 2023/51/B/ST6/00545.
 \end{acknowledgments}

\bibliography{ecai2024-haii}

@Article{falldetection2023,
AUTHOR = {Newaz, Nishat Tasnim and Hanada, Eisuke},
TITLE = {The Methods of Fall Detection: A Literature Review},
JOURNAL = {Sensors},
VOLUME = {23},
YEAR = {2023},
NUMBER = {11},
ARTICLE-NUMBER = {5212},
URL = {https://www.mdpi.com/1424-8220/23/11/5212},
PubMedID = {37299939},
ISSN = {1424-8220},
DOI = {10.3390/s23115212}
}

@InProceedings{pietraszek2004intrusion,
author="Pietraszek, Tadeusz",
editor="Jonsson, Erland
and Valdes, Alfonso
and Almgren, Magnus",
title="Using Adaptive Alert Classification to Reduce False Positives in Intrusion Detection",
booktitle="Recent Advances in Intrusion Detection",
year="2004",
publisher="Springer Berlin Heidelberg",
address="Berlin, Heidelberg",
pages="102--124",
isbn="978-3-540-30143-1"
}

@InProceedings{gama2004drift,
author="Gama, Jo{\~a}o
and Medas, Pedro
and Castillo, Gladys
and Rodrigues, Pedro",
editor="Bazzan, Ana L. C.
and Labidi, Sofiane",
title="Learning with Drift Detection",
booktitle="Advances in Artificial Intelligence -- SBIA 2004",
year="2004",
publisher="Springer Berlin Heidelberg",
address="Berlin, Heidelberg",
pages="286--295",
isbn="978-3-540-28645-5"
}

@article{karimian2024dadrift,
title = {Concept drift handling: A domain adaptation perspective},
journal = {Expert Systems with Applications},
volume = {224},
pages = {119946},
year = {2023},
issn = {0957-4174},
doi = {https://doi.org/10.1016/j.eswa.2023.119946},
url = {https://www.sciencedirect.com/science/article/pii/S0957417423004487},
author = {Mahmood Karimian and Hamid Beigy}
}

@inproceedings{kuk2023shaprootcause,
author = {Kuk, Micha\l{} and Bobek, Szymon and Veloso, Bruno and Rajaoarisoa, Lala and Nalepa, Grzegorz J.},
title = {Feature Importances as Tool for Root Cause Analysis in Time-Series Events},
year = {2023},
isbn = {978-3-031-36029-9},
publisher = {Springer-Verlag},
address = {Berlin, Heidelberg},
url = {https://doi.org/10.1007/978-3-031-36030-5_33},
doi = {10.1007/978-3-031-36030-5_33},
booktitle = {Computational Science – ICCS 2023: 23rd International Conference, Prague, Czech Republic, July 3–5, 2023, Proceedings, Part V},
pages = {408–416},
numpages = {9},
location = {<conf-loc content-type="InPerson">Prague, Czech Republic</conf-loc>}
}

@inproceedings{10.1007/978-3-031-36027-5_37, author = {Jakubowski, Jakub and Stanisz, Przemys\l{}aw and Bobek, Szymon and Nalepa, Grzegorz J.}, title = {Towards Online Anomaly Detection in Steel Manufacturing Process}, year = {2023}, isbn = {978-3-031-36026-8}, publisher = {Springer-Verlag}, address = {Berlin, Heidelberg}, url = {https://doi.org/10.1007/978-3-031-36027-5_37}, doi = {10.1007/978-3-031-36027-5_37}, booktitle = {Computational Science – ICCS 2023: 23rd International Conference, Prague, Czech Republic, July 3–5, 2023, Proceedings, Part IV}, pages = {469–482}, numpages = {14}, location = {Prague, Czech Republic} }

@article{10.1145/335191.335388, author = {Breunig, Markus M. and Kriegel, Hans-Peter and Ng, Raymond T. and Sander, J\"{o}rg}, title = {LOF: identifying density-based local outliers}, year = {2000}, issue_date = {June 2000}, publisher = {Association for Computing Machinery}, address = {New York, NY, USA}, volume = {29}, number = {2}, issn = {0163-5808}, url = {https://doi.org/10.1145/335191.335388}, doi = {10.1145/335191.335388}, journal = {SIGMOD Rec.}, month = {may}, pages = {93–104}, numpages = {12} }

@article{SHIN2005395,
title = {One-class support vector machines—an application in machine fault detection and classification},
journal = {Computers \& Industrial Engineering},
volume = {48},
number = {2},
pages = {395-408},
year = {2005},
issn = {0360-8352},
doi = {https://doi.org/10.1016/j.cie.2005.01.009},
url = {https://www.sciencedirect.com/science/article/pii/S0360835205000100},
author = {Hyun Joon Shin and Dong-Hwan Eom and Sung-Shick Kim}
}

@inproceedings{Rumelhart1986LearningIR,
  title={Learning internal representations by error propagation},
  author={David E. Rumelhart and Geoffrey E. Hinton and Ronald J. Williams},
  year={1986},
  url={https://api.semanticscholar.org/CorpusID:62245742}
}

@misc{motiian2017unified,
      title={Unified Deep Supervised Domain Adaptation and Generalization}, 
      author={Saeid Motiian and Marco Piccirilli and Donald A. Adjeroh and Gianfranco Doretto},
      year={2017},
      eprint={1709.10190},
      archivePrefix={arXiv},
      primaryClass={cs.CV}
}

@article{10.1093/biomet/41.1-2.100,
    author = {PAGE, E. S.},
    title = "{Continuous inspection schemes}",
    journal = {Biometrika},
    volume = {41},
    number = {1-2},
    pages = {100-115},
    year = {1954},
    month = {06},
    issn = {0006-3444},
    doi = {10.1093/biomet/41.1-2.100},
    url = {https://doi.org/10.1093/biomet/41.1-2.100}
    
}

@INPROCEEDINGS{4595271,
  author={Perez-Cruz, Fernando},
  booktitle={2008 IEEE International Symposium on Information Theory}, 
  title={Kullback-Leibler divergence estimation of continuous distributions}, 
  year={2008},
  volume={},
  number={},
  pages={1666-1670},
  doi={10.1109/ISIT.2008.4595271}}

@inproceedings{NIPS2017_8a20a862,
 author = {Lundberg, Scott M and Lee, Su-In},
 booktitle = {Advances in Neural Information Processing Systems},
 editor = {I. Guyon and U. Von Luxburg and S. Bengio and H. Wallach and R. Fergus and S. Vishwanathan and R. Garnett},
 pages = {},
 publisher = {Curran Associates, Inc.},
 title = {A Unified Approach to Interpreting Model Predictions},
 url = {https://proceedings.neurips.cc/paper_files/paper/2017/file/8a20a8621978632d76c43dfd28b67767-Paper.pdf},
 volume = {30},
 year = {2017}
}

@article{10.1145/1541880.1541882,
author = {Chandola, Varun and Banerjee, Arindam and Kumar, Vipin},
title = {Anomaly detection: A survey},
year = {2009},
issue_date = {July 2009},
publisher = {Association for Computing Machinery},
address = {New York, NY, USA},
volume = {41},
number = {3},
issn = {0360-0300},
url = {https://doi.org/10.1145/1541880.1541882},
doi = {10.1145/1541880.1541882},
journal = {ACM Comput. Surv.},
month = {jul},
articleno = {15},
numpages = {58}
}

@INPROCEEDINGS{9564228,
  author={Jakubowski, Jakub and Stanisz, Przemysław and Bobek, Szymon and Nalepa, Grzegorz J.},
  booktitle={2021 IEEE 8th International Conference on Data Science and Advanced Analytics (DSAA)}, 
  title={Explainable anomaly detection for Hot-rolling industrial process}, 
  year={2021},
  volume={},
  number={},
  pages={1-10},
  doi={10.1109/DSAA53316.2021.9564228}}

@INPROCEEDINGS{9892939,
  author={Gerz, Fabian and Bastürk, Tolga Renan and Kirchhoff, Julian and Denker, Joachim and Al-Shrouf, Loui and Jelali, Mohieddine},
  booktitle={2022 International Joint Conference on Neural Networks (IJCNN)}, 
  title={A Comparative Study and a New Industrial Platform for Decentralized Anomaly Detection Using Machine Learning Algorithms}, 
  year={2022},
  volume={},
  number={},
  pages={1-8},
  doi={10.1109/IJCNN55064.2022.9892939}}

@book{molnar2020interpretable,
  title={Interpretable Machine Learning},
  author={Molnar, Christoph},
  year={2020},
  publisher={Lulu.com}
}

@misc{ribeiro2016why,
      title={"Why Should I Trust You?": Explaining the Predictions of Any Classifier}, 
      author={Marco Tulio Ribeiro and Sameer Singh and Carlos Guestrin},
      year={2016},
      eprint={1602.04938},
      archivePrefix={arXiv},
      primaryClass={cs.LG}
}

@InProceedings{10.1007/978-3-030-77980-1_34,
author="Bobek, Szymon
and Nalepa, Grzegorz J.",
editor="Paszynski, Maciej
and Kranzlm{\"u}ller, Dieter
and Krzhizhanovskaya, Valeria V.
and Dongarra, Jack J.
and Sloot, Peter M. A.",
title="Introducing Uncertainty into Explainable AI Methods",
booktitle="Computational Science -- ICCS 2021",
year="2021",
publisher="Springer International Publishing",
address="Cham",
pages="444--457",
isbn="978-3-030-77980-1"
}

@misc{wachter2018counterfactual,
      title={Counterfactual Explanations without Opening the Black Box: Automated Decisions and the GDPR}, 
      author={Sandra Wachter and Brent Mittelstadt and Chris Russell},
      year={2018},
      eprint={1711.00399},
      archivePrefix={arXiv},
      primaryClass={cs.AI}
}

@InProceedings{10.1007/978-3-031-50396-2_5,
author="Jakubowski, Jakub
and Stanisz, Przemys{\l}aw
and Bobek, Szymon
and Nalepa, Grzegorz J.",
editor="Nowaczyk, S{\l}awomir
and Biecek, Przemys{\l}aw
and Chung, Neo Christopher
and Vallati, Mauro
and Skruch, Pawe{\l}
and Jaworek-Korjakowska, Joanna
and Parkinson, Simon
and Nikitas, Alexandros
and Atzm{\"u}ller, Martin
and Kliegr, Tom{\'a}{\v{s}}
and Schmid, Ute
and Bobek, Szymon
and Lavrac, Nada
and Peeters, Marieke
and van Dierendonck, Roland
and Robben, Saskia
and Mercier-Laurent, Eunika
and Kayakutlu, G{\"u}lg{\"u}n
and Owoc, Mieczyslaw Lech
and Mason, Karl
and Wahid, Abdul
and Bruno, Pierangela
and Calimeri, Francesco
and Cauteruccio, Francesco
and Terracina, Giorgio
and Wolter, Diedrich
and Leidner, Jochen L.
and Kohlhase, Michael
and Dimitrova, Vania",
title="Explainable Anomaly Detection in Industrial Streams",
booktitle="Artificial Intelligence. ECAI 2023 International Workshops",
year="2024",
publisher="Springer Nature Switzerland",
address="Cham",
pages="87--100",
isbn="978-3-031-50396-2"
}

@article{DING2023144,
title = {A high-precision and transparent step-wise diagnostic framework for hot-rolled strip crown},
journal = {Journal of Manufacturing Systems},
volume = {71},
pages = {144-157},
year = {2023},
issn = {0278-6125},
doi = {https://doi.org/10.1016/j.jmsy.2023.09.007},
url = {https://www.sciencedirect.com/science/article/pii/S0278612523001899},
author = {Chengyan Ding and Jie Sun and Xiaojian Li and Wen Peng and Dianhua Zhang}
}

@misc{jakubowski2024artificial,
      title={Artificial Intelligence Approaches for Predictive Maintenance in the Steel Industry: A Survey}, 
      author={Jakub Jakubowski and Natalia Wojak-Strzelecka and Rita P. Ribeiro and Sepideh Pashami and Szymon Bobek and Joao Gama and Grzegorz J Nalepa},
      year={2024},
      eprint={2405.12785},
      archivePrefix={arXiv},
      primaryClass={cs.AI}
}

@inproceedings{BaenaGarc2005EarlyDD,
  title={Early Drift Detection Method},
  author={Manuel Baena-Garc and Jos{\'e} Avila and Albert Bifet and Ricard Gavald and Rafael Morales-Bueno},
  year={2005},
  url={https://api.semanticscholar.org/CorpusID:15672006}
}

@inproceedings{bifet2007learning,
  title={Learning from time-changing data with adaptive windowing},
  author={Bifet, Albert and Gavalda, Ricard},
  booktitle={Proceedings of the 2007 SIAM international conference on data mining},
  pages={443--448},
  year={2007},
  organization={SIAM}
}

@article{AGRAHARI20229523,
title = {Concept Drift Detection in Data Stream Mining : A literature review},
journal = {Journal of King Saud University - Computer and Information Sciences},
volume = {34},
number = {10, Part B},
pages = {9523-9540},
year = {2022},
issn = {1319-1578},
doi = {https://doi.org/10.1016/j.jksuci.2021.11.006},
url = {https://www.sciencedirect.com/science/article/pii/S1319157821003062},
author = {Supriya Agrahari and Anil Kumar Singh},

}

@article{article,
author = {Gama, João and Aguilar-Ruiz, Jesús and Klinkenberg, Ralf},
year = {2008},
month = {05},
pages = {251-252},
title = {Knowledge discovery from data streams},
volume = {12},
journal = {Intell. Data Anal.},
doi = {10.3233/IDA-2008-12301}
}

@inbook{stefanowski2017stream,
  title={Stream Classification.},
  author={Stefanowski, Jerzy and Brzezinski, Dariusz},
  booktitle = {Encyclopedia of Machine learning},
  year = {2017},
  publisher = {Springer},
  pages = {1191 - 1195}
}

@article{brzezinski2021impact,
  title={The impact of data difficulty factors on classification of imbalanced and concept drifting data streams},
  author={Brzezinski, Dariusz and Minku, Leandro L and Pewinski, Tomasz and Stefanowski, Jerzy and Szumaczuk, Artur},
  journal={Knowledge and Information Systems},
  volume={63},
  number={6},
  pages={1429--1469},
  year={2021},
  publisher={Springer}
}

@article{stepka2024multi,
  title={A Multi--Criteria Approach for Selecting an Explanation from the Set of Counterfactuals Produced by an Ensemble of Explainers},
  author={Stepka, Ignacy and Lango, Mateusz and Stefanowski, Jerzy},
  journal={International Journal of Applied Mathematics and Computer Science},
  volume={34},
  number={1},
  year = {2024},
  pages={119--133}
}

@article{krawczyk2017ensemble,
  title={Ensemble learning for data stream analysis: A survey},
  author={Krawczyk, Bartosz and Minku, Leandro L and Gama, Jo{\~a}o and Stefanowski, Jerzy and Wo{\'z}niak, Micha{\l}},
  journal={Information Fusion},
  volume={37},
  pages={132--156},
  year={2017},
  publisher={Elsevier}
}

@article{guidotti2018survey,
  title={A survey of methods for explaining black box models},
  author={Guidotti, Riccardo and Monreale, Anna and Ruggieri, Salvatore and Turini, Franco and Giannotti, Fosca and Pedreschi, Dino},
  journal={ACM computing surveys (CSUR)},
  volume={51},
  number={5},
  pages={1--42},
  year={2018},
  publisher={ACM New York, NY, USA}
}

@article{Aleskerov1997CARDWATCHAN,
  title={CARDWATCH: a neural network based database mining system for credit card fraud detection},
  author={Emin Aleskerov and Bernd Freisleben and R. Bharat Rao},
  journal={Proceedings of the IEEE/IAFE 1997 Computational Intelligence for Financial Engineering (CIFEr)},
  year={1997},
  pages={220-226},
  url={https://api.semanticscholar.org/CorpusID:13509083}
}

@inproceedings{Fujimaki2005AnAT,
  title={An approach to spacecraft anomaly detection problem using kernel feature space},
  author={Ryohei Fujimaki and Takehisa Yairi and Kazuo Machida},
  booktitle={Knowledge Discovery and Data Mining},
  year={2005},
  url={https://api.semanticscholar.org/CorpusID:1735180}
}

@misc{ganin2016domainadversarial,
      title={Domain-Adversarial Training of Neural Networks}, 
      author={Yaroslav Ganin and Evgeniya Ustinova and Hana Ajakan and Pascal Germain and Hugo Larochelle and François Laviolette and Mario Marchand and Victor Lempitsky},
      year={2016},
      eprint={1505.07818},
      archivePrefix={arXiv},
      primaryClass={stat.ML}
}

\end{document}